\documentclass[10pt,twocolumn,letterpaper]{article}

\usepackage[pagenumbers]{iccv} %

\usepackage{gensymb}
\usepackage{colortbl}

\usepackage{xcolor}
\usepackage[textsize=tiny]{todonotes}
\usepackage{ifthen}
\newboolean{comments}
\setboolean{comments}{true}  %

\definecolor{iccvblue}{rgb}{0.21,0.49,0.74}
\usepackage[pagebackref,breaklinks,colorlinks,allcolors=iccvblue]{hyperref}
\usepackage{multirow}
        
\def\paperID{*****} %
\def\confName{ICCV}
\def\confYear{2025}

\title{EscherNet++: Simultaneous Amodal Completion and Scalable View Synthesis through Masked Fine-Tuning and Enhanced Feed-Forward 3D Reconstruction}

\author{
Xinan Zhang$^{1}$ \quad Muhammad Zubair Irshad$^{2}$ \quad Anthony Yezzi$^{1}$ \quad Yi-Chang Tsai$^{1}$ \quad Zsolt Kira$^{1}$\\
Georgia Institute of Technology$^{1}$ \quad Toyota Research Institute$^{2}$\\
}

\begin{document}

\twocolumn[{%
\renewcommand\twocolumn[1][]{#1}%
\maketitle
\begin{center}
    \centering\includegraphics[width=18cm]{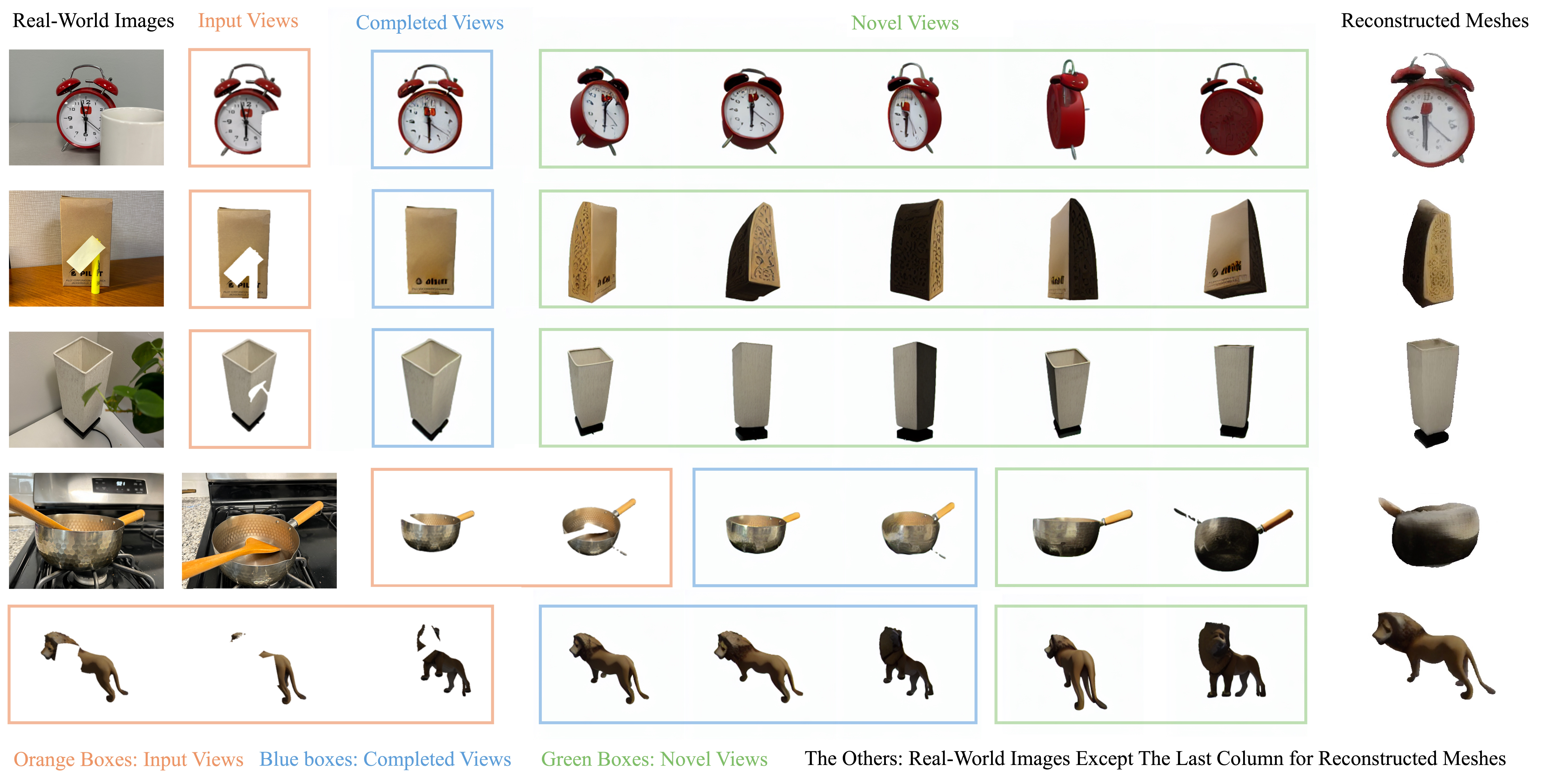}\captionsetup{width=\linewidth}
    \captionof{figure}{
Given occluded input views, \textbf{EscherNet++} simultaneously completes occluded views and synthesizes novel ones—without requiring multiple specialized models. Synthesized views can be queried from any viewpoint, enabling seamless integration with feed-forward 3D reconstruction models. We empirically estimate camera poses, use Entity V2 \cite{qi2022high} for segmentation, and InstantMesh \cite{xu2024instantmesh} for fast mesh reconstruction, completing the process in about 2 minutes in this demo. Our method generalizes to unseen real-world data, with the last row showcasing samples from our synthesized occlusion benchmark, \textbf{OccNVS}, unseen during training.}
   \label{fig:teaser}
\end{center}%
}]

\begin{abstract}

We propose EscherNet++, a masked fine-tuned diffusion model that can synthesize novel views of objects in a zero-shot manner with amodal completion ability. Existing approaches utilize multiple stages and complex pipelines to first hallucinate missing parts of the image and then perform novel view synthesis, which fail to consider cross-view dependencies and require redundant storage and computing for separate stages. Instead, we apply masked fine-tuning including input-level and feature-level masking to enable an end-to-end model with the improved ability to synthesize novel views and conduct amodal completion.
In addition, we empirically integrate our model with other feed-forward image-to-mesh models without extra training and achieve competitive results with reconstruction time decreased by 95\%, thanks to its ability to synthesize arbitrary query views. Our method's scalable nature further enhances fast 3D reconstruction. Despite fine-tuning on a smaller dataset and batch size, our method achieves state-of-the-art results, improving PSNR by 3.9 and Volume IoU by 0.28 on occluded tasks in 10-input settings, while also generalizing to real-world occluded reconstruction.
\end{abstract}
    
\section{Introduction}
\label{sec:intro}

\begin{figure*}
  \centering
  \resizebox{1.0\textwidth}{!}{%
  \includegraphics[width=18cm]{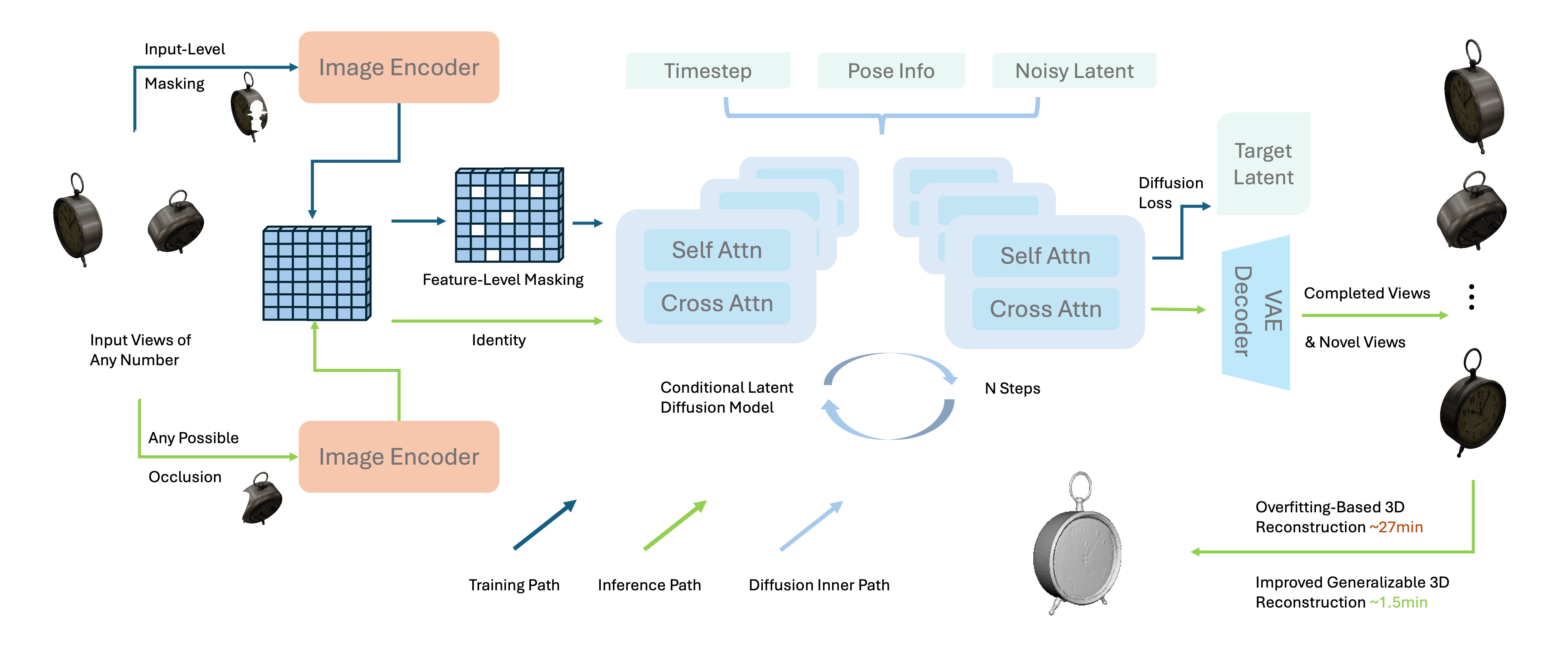}

  }
\caption{\textbf{The pipeline of EscherNet++}. Our unified model enables simultaneous novel view synthesis and amodal completion. During training, hierarchical masking—at both input and feature levels—helps the model learn complete geometry from occluded views while improving robustness. 
During inference, our model not only supports commonly used overfitting approaches—such as NueS ~\cite{wang2021neus}, which iteratively refines geometry—but also seamlessly integrates with pre-trained feed-forward models like InstantMesh ~\cite{xu2024instantmesh}. We empirically find that this integration achieves competitive performance while significantly reducing computational time.
} %
\label{fig:pipeline}
\end{figure*}

Novel view synthesis (NVS) of objects 
is an important topic in computer vision due to its wide range of applications, including virtual and augmented reality~\cite{kuang2022neroic, nguyen2024semantically}, computer graphics~\cite{instructnerf2023}, robotics~\cite{irshad2024neuralfieldsroboticssurvey, zhu20243dgaussiansplattingrobotics} and 3D reconstruction~\cite{irshad2022shapo, li2023neuralangelo, lunayach2023fsd} . It involves generating new images of 
an object from viewpoints that were not observed during data capture, enabling more immersive and interactive experiences. The computer vision community has witnessed fast progress in this field. One of the recent prominent approaches is neural radiance fields (NeRF)~\cite{mildenhall2021nerf}, which achieves high-quality results by modeling the scene as a continuous volumetric function using a neural network. NeRF and its following works have revolutionized novel view synthesis but come with several limitations that hinder their practical application, including 1) slow training/rendering speeds, 2) limited extrapolation/few-shot/generalization ability and 3) inability to handle occlusion well.

Various methods have been proposed to alleviate these problems while maintaining the quality of the synthesis. Grid-based methods such as Instant-NGP~\cite{muller2022instant} and point-based methods such as 3D Gaussian splatting (3DGS)~\cite{kerbl20233d} have reduced training and rendering time. In addition, techniques have been proposed to incorporate learned prior knowledge to help improve generalization, and few-shot abilities~\cite{yu2021pixelnerf, irshad2023neo}. Diffusion methods~\cite{ho2020denoising, song2020denoising, rombach2022high}, which are a group of generative models previously used in content generation, began to gain popularity in NVS. These diffusion-based NVS models~\cite{liu2023zero, shi2023zero123++, liu2023syncdreamer, huang2024epidiff, kant2024spad, voleti2024sv3d, long2024wonder3d, zheng2024free3d, kong2024eschernet, xu2024sparp, tang2024mvdiffusion++} leverage learned priors from large-scale datasets, allowing them to generate plausible novel views even with sparse input data or extreme viewpoint extrapolation. 

Although diffusion-based methods usually require large-scale training, they tend to have strong generalization capabilities. Among this group, most works only support conditioning on one single image; EscherNet~\cite{kong2024eschernet} stands out for its ability to generate high-quality consistent views and support multiple inputs as the condition. As our base model, it utilizes self and cross attention mechanisms to ensure consistency among views during diffusion. Besides, occlusion in input views can be a challenge. Diffusion models have also been used in amodal completion to deal with occlusion~\cite{ozguroglu2024pix2gestalt, Dogaru2024Gen3DSR, brooks2023instructpix2pix}, completing the occluded inputs for the subsequent tasks such as NVS; however, current amodal completion models typically serve as a stand-alone model and primarily focus on single-view context, while neglecting multi-view constraints and references.

Departing from existing approaches that often treat these tasks separately~\cite{Dogaru2024Gen3DSR, ozguroglu2024pix2gestalt}, sometimes leading to inefficiencies in inference and inconsistencies in geometric and appearance modeling, we ask \textit{"Can these two problems be solved with a  more integrated solution?"} Such a solution should be able to 1) leverage a shared understanding of object semantics and geometry from the input views with possible occlusions, which requires a unified dataset supporting learning on NVS and amodal completion and 2) possess the ability to be optimized collectively for both tasks, which requires universal training mechanisms for improving robustness. The first requirement leads us to curate a dataset in order to simulate possible occlusions on objects from the input views, which we term as input-level masking. Inspired by recent success achieved by masked training~\cite{he2022masked, woo2023convnext}, we continue to apply feature-level masking on the encoded feature maps as well. Experiments showcase that input-level masking is the key to robustness to occlusions in input views, while feature-level masking can further enhance the performance by better understanding of semantics and geometry, as we show in Sec. \ref{sec:exp1}. Together, these two masking approaches form the masked fine-tuning mechanism proposed by our work {\bf EscherNet++}, a unified model adapted from EscherNet, which simultaneously enables amodal completion and NVS conditioned on a flexible number of input views and also supports synchronically multi-view synthesis of a flexible number, as shown in Fig \ref{fig:pipeline}.

Moreover, our framework seamlessly integrates with other fast, feed-forward 3D reconstruction models due to its ability to synthesize views from any given query viewpoint and across various numbers of views. We demonstrate this ability with InstantMesh~\cite{xu2024instantmesh}, a recent fast 3D reconstruction network. Additionally, the performance can be improved to a level on par with SoTA overfitting methods such as NeuS~\cite{wang2021neus}, by querying our model from extra viewpoints and providing synthesized views to InstantMesh. No additional training or significantly longer inference time is needed because of the scable nature of our model. Extensive experiments are conducted on occluded zero-shot NVS and 3D reconstruction benchmarks, showing that our model achieves outstanding performance in the above tasks. We term the benchmark {\bf OccNVS}, which is composed of occluded views and complete views from~\cite{downs2022google, tremblay2022rtmv, mildenhall2021nerf}. In summary, our contributions are as follows:
\begin{itemize}
    \item We propose a unified diffusion-based network {\bf EscherNet++} as shown in Fig. \ref{fig:pipeline}, designed for \textbf{occlusion-aware novel view synthesis}. It flexibly adapts to varying numbers of input and output views, extending the original task for \textbf{multi-view amodal completion}—a challenging yet underexplored task. 
    \item  Introduce an effective approach to~\textbf{enhance fast 3D reconstruction using pre-trained feed-forward models}, leveraging the scalability and consistency of our synthesized novel views without requiring additional fine-tuning
    \item Our proposed work {excels in extensive experiments} on NVS and 3D reconstruction, particularly \textbf{under occlusions}, outperforming prior work by an average of {3.9 PSNR in occluded NVS tests and 0.28 Volume IoU} in occluded 3D reconstruction tests with 10-input settings.
\end{itemize}

\section{Related Work}
\label{sec:related_work}

Neural Radiance Fields~\cite{mildenhall2021nerf} and its variants~\cite{muller2022instant, wang2021neus, yu2021pixelnerf} have gained significant attention in novel view synthesis for their photorealistic results. However, their high computational cost, long training times, and implicit nature limit applicability in scenarios requiring efficiency. To address these limitations, methods such as 3D Gaussian Splatting (3DGS)~\cite{kerbl20233d} adopt point-based representations for efficient 3D scene modeling. Advancements in NeRF and 3DGS have enabled various downstream applications, including multi-view consistent segmentation and editing by propagating semantics across views~\cite{kerr2023lerf, siddiqui2023panoptic, yu2025pogs, ye2024gaussian}, as well as dynamic scene modeling by incorporating time as a dimension~\cite{li2021neural, wu20244d}. While successful, generalizability remains a challenge. Efforts to address this~\cite{yu2021pixelnerf, irshad2023neo, charatan2024pixelsplat} continue, alongside emerging alternatives such as diffusion models.

\textbf{Diffusion Model for Novel View Synthesis:} Diffusion models have emerged as a promising alternative for novel view synthesis (NVS), addressing the extrapolation and input limitations of NeRF and Gaussian Splatting. Zero-1-to-3~\cite{liu2023zero}, trained on Objaverse~\cite{deitke2023objaverse}, repurposes a pre-trained text-to-image diffusion model to generate novel views from arbitrary viewpoints using a single input image. Building on this, subsequent works enhance single-image-to-3D synthesis while incorporating additional functionalities, often leveraging attention mechanisms for improved view consistency~\cite{liu2023syncdreamer, shi2023zero123++, zheng2024free3d, long2024wonder3d, huang2024epidiff, kant2024spad}. EscherNet~\cite{kong2024eschernet} extends Zero-1-to-3 with cross- and self-attention for reference-to-target consistency and supports conditioning on multiple input images, making it well-suited for integration with 3D reconstruction models. Beyond NVS, diffusion-based methods naturally extend to 3D reconstruction, following either a two-stage or single-stage approach. Two-stage methods synthesize novel views as intermediate inputs to 3D reconstruction networks, which can be overfitting-based or generalizable~\cite{liu2023syncdreamer, wang2021neus, liu2024one, xu2024instantmesh, wang2023imagedream}. In contrast, single-stage methods directly infer 3D geometry from 2D observations without an explicit NVS step, often relying on implicit neural representations or volumetric predictions~\cite{tang2023dreamgaussian, zou2024triplane, hong2023lrm}. However, these approaches currently struggle to match the performance of two-stage pipelines.

\textbf{Amodel Completion:} focuses on predicting and filling in occluded object regions to ensure a coherent understanding of object geometry and appearance. Recently, diffusion models have been explored for this task due to their ability to generate high-quality and realistic completions. Gen3DSR~\cite{Dogaru2024Gen3DSR} fine-tunes a stable diffusion model~\cite{rombach2022high} following InstructPix2Pix~\cite{brooks2023instructpix2pix} to complete segmented objects in a scene. Similarly,~\cite{xu2024amodal} applies a pre-trained diffusion inpainting model~\cite{rombach2022high} to iteratively reconstruct occluded objects while removing the background.~\cite{zhan2024amodal} explores both one-stage and two-stage diffusion models for amodal mask prediction, while Pix2Gestalt~\cite{ozguroglu2024pix2gestalt} conditions completion on user prompts using VAE~\cite{kingma2013auto} and CLIP~\cite{radford2021learning} embeddings.

\textbf{Masked Training for Diffusion models:} Masked training originated from works like Masked Autoencoders (MAE)~\cite{he2022masked, pang2022masked, irshad2024nerfmae}, a technique where a large portion of input data is masked, and the model learns to reconstruct the missing information. This method is particularly effective for self-supervised learning, helping models understand the underlying structure of data without explicit labels. More recently, researchers have started integrating masked training into diffusion models. It has been applied to enhance the models' ability to contextual relation learning among object semantic parts~\cite{gao2023mdtv2}, allow for future/past prediction~\cite{voleti2022mcvd}, achieve a balance between sample quality and diversity~\cite{ho2022classifier}, etc. In addition, Diffusion based amodal completion models also rely on training with masked inputs~\cite{Dogaru2024Gen3DSR, xu2024amodal, zhan2024amodal, ozguroglu2024pix2gestalt}.

\section{Methodology}
NVS becomes significantly more challenging when reference views contain occlusions. To address this, occluded regions should be handled through amodal completion, enabling NVS to be performed on completed views without being affected by missing information. Rather than constructing a pipeline of separate, specialized models, we propose a unified approach that simultaneously facilitates both NVS and amodal completion.
This design offers several advantages. First, it has smaller storage footprint, enabling faster processing time. Second, it considers the problem holistically with multi-view reference taken into account naturally, unifying multi-view amodal completion and NVS tasks.
To address these requirements, we introduce~\textbf{EscherNet++}, detailed in this section. We first provide background in Sec. \ref{sec:bg}, followed by our masked fine-tuning approach in Sec.~\ref{sec:maskedtraining}, and our view-to-3D reconstruction method in Sec.~\ref{sec:nvs23d}. An overview of the pipeline is illustrated in Fig.~\ref{fig:teaser}.

\subsection{Background}
\label{sec:bg}

We begin by introducing key concepts related to diffusion models, including their general working principles. Additionally, we discuss a specialized application in NVS enabled by EscherNet~\cite{kong2024eschernet}.

\subsubsection{Diffusion model}

Diffusion models have gained popularity as generative models that learn data distributions by iteratively denoising a signal in inference that is usually corrupted by Gaussian noise~\cite{ho2020denoising, song2020denoising, rombach2022high}. The core idea involves a Markovian forward process that gradually adds noise to the data and a reverse process that learns to denoise and generate realistic samples. 
In the forward process, gaussian noise is gradually added to a original data sample.
In the reverse process, the model aims to denoise the corrupted data given the current timestep. Based on this, subsequent works continue to improve its practicability~\cite{song2020denoising, rombach2022high}.

\subsubsection{EscherNet}

EscherNet is the first work designed to incorporate multiple input views in diffsion-based NVS. It realizes generalizable zero-shot novel view synthesis, with support for dynamic input views. The general goal of EscherNet can be formulated as a conditional generation problem:
\begin{equation} 
X^T \sim p(X^T \mid X^R, P^R, P^T)
\label{eq:eschernet1}
\end{equation}
\noindent where $X^T$, $P^T$ represent target novel views and query poses, and $T^R$, $P^R$ are input reference views and related poses. Different from previous works, the number of input and target views is not restricted by EscherNet.
and can vary depending on the specific use cases.
It makes use of an existing latent diffusion model~\cite{rombach2022high}, pre-trained on web-scale data, with U-Net~\cite{ronneberger2015u} composed of residual blocks~\cite{he2016deep} and transformer blocks~\cite{vaswani2017attention} as the backbone. EscherNet conditions its generation process on input visual information from a lightweight CNN-based vision encoder~\cite{woo2023convnext}, and pose information from its camera positional encoding (CaPE). Low-resolution latents are decoded to images~\cite{kingma2013auto} as in~\cite{rombach2022high}. It is trained on rendered Objaverse 1.0 dataset~\cite{liu2023zero} and achieves SoTA multi-view NVS results.
However, its inability to handle occlusion in input views and time-consuming %
image-to-3D reconstruction process still persist as main challenges. Therefore, we propose two corresponding techniques in this paper to mitigate these challenges i.e. 1) Masked fine-tuning on EscherNet with input-level and feature-level masking to allow for NVS with occluded inputs. 2) Training-free integration with feed-forward 3D reconstruction methods which cuts down the reconstruction time by 95\% while maintaining performance.

\begin{figure}
  \centering
  \resizebox{0.48\textwidth}{!}{\includegraphics[width=18cm]{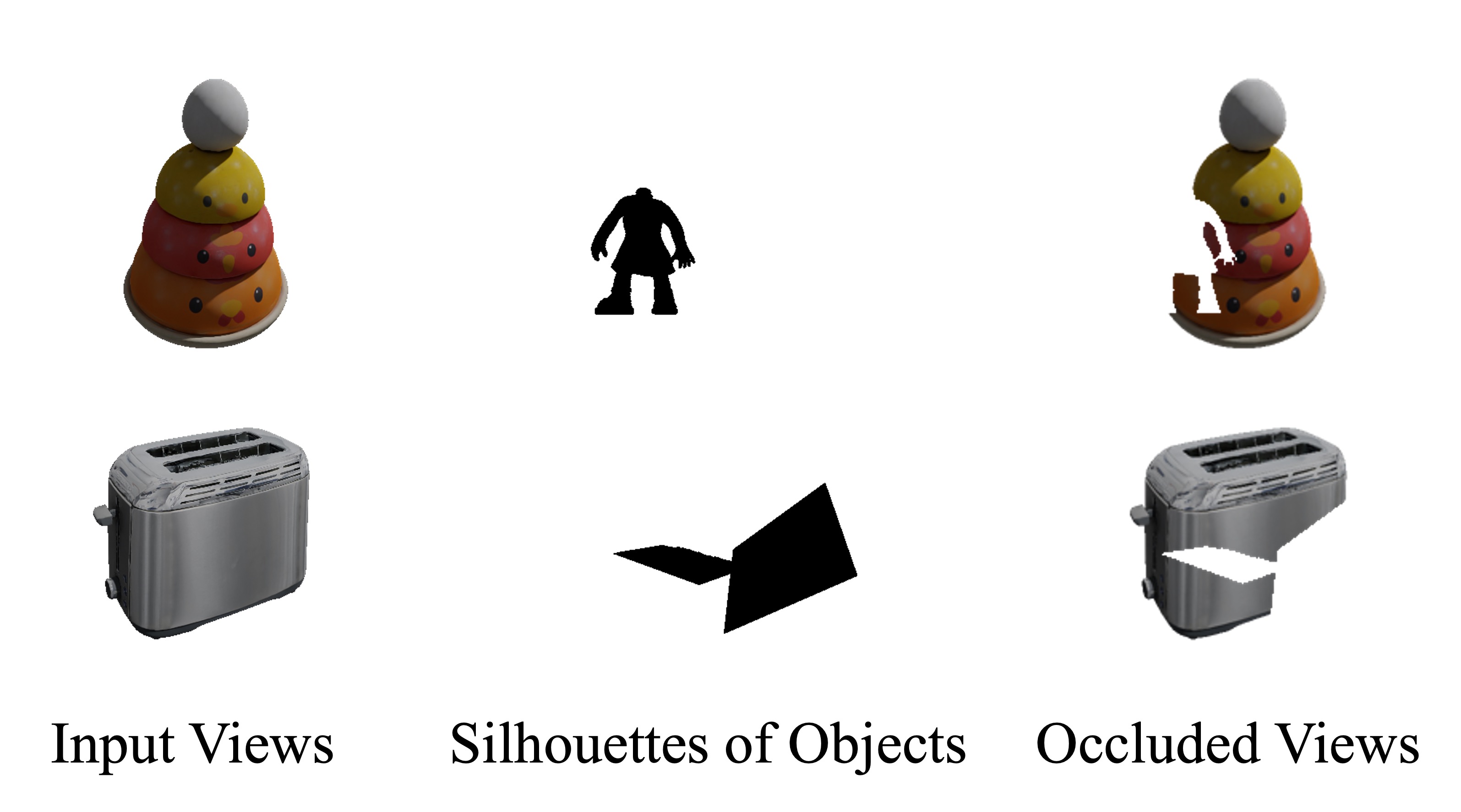}
  }
\caption{Examples of how input-level masking is applied. Silhouettes are extracted from rendered objects and overlayed on complete input views to get occluded views paired with groundtruth.}
\label{fig:masking}
\end{figure}

\subsection{Masked Fine-Tuning}
\label{sec:maskedtraining}

Built upon EscherNet, we devise a hierarchical masked fine-tuning method in order to deal with possible occlusions in input views. Contrary to previous attempts to handle occlusions separately from NVS, we aim to achieve an end-to-end model that can synthesize novel views and complete the occluded regions in input views simultaneously. There are two key aspects to consider when tackling the compound problem, 1) dataset acquisition and 2) training method.  

\textbf{Curated Dataset:} A well-structured dataset is crucial for training a model to handle the problem effectively. To achieve this, we need a dataset containing both occluded and complete versions of the same samples. There have been several examples available, like occluded SA-1B dataset  ~\cite{kirillov2023segment} processed by  ~\cite{ozguroglu2024pix2gestalt}. However, it should also support training for novel view synthesis of various objects to fit the original design of EscherNet.

The requirements on the dataset lead us to create a paired dataset curated from Objaverse-1.0  ~\cite{deitke2023objaverse}.%
We employ silhouettes of objects as masks to randomly overlay occlusions onto objects in the dataset, as shown in Fig. \ref{fig:pipeline} \& \ref{fig:masking}. %
Specifically, we sampled single objects from then rendered Objaverse dataset to extract their silhouettes. Then we group, rescale, shift them to create various occlusions. Finally, these occlusions are randomly overlayed onto complete objects in the rendered Objaverse dataset with a certain probability to simulate any possible occlusions. In our experiments, each input view in the original dataset has 50 percent chance to be occluded, while the model is trained to predict the complete views from query poses. %
By applying this technique consistently across the dataset, we curate a paired dataset that maintains the same size as the original dataset while incorporating diverse occlusion patterns.

\textbf{Input-Level \& Feature-Level Masking:} We fine-tune the model using two techniques, input-level masking and feature-level masking. Input level masking can be achieved with the above curated dataset naturally. Similar to the original training of Eschernet, we randomly choose three input views with 50 percent chance of being partially occluded, the model learns to synthesize novel three other complete views. In addition, inspired by previous works  ~\cite{he2022masked, woo2023convnext, irshad2024nerfmae, gao2023mdtv2}, we propose to further  randomly mask the encoded input feature maps to further improve the performance, as shown in Sec. \ref{sec:exp1}. We hypothesize that proper feature-leveling masking can improve model's ability in overall comprehension of the object in two ways--better understanding of semantics and better capture of intricate structure details. Specifically, each input image has a certain chance of being chosen for feature-level masking. Once picked, feature vectors in the feature map will be randomly masked out with 50 percent chance. We empirically found that 25 percent is a suitable choice for feature-level masking probability as shown in App. \ref{app:exp1}. That is, around 1/4 of input data will be processed by random feature-level masking during training. %

\subsection{Novel View Synthesis to 3D reconstruction}
\label{sec:nvs23d}

Reconstructing objects from synthesized novel views is a crucial downstream task. In this section, we aim to combine the model from the last section with other SoTA image-to-3D methods for fast and high-quality object reconstruction. Broadly, two main approaches exist: 1) Overfitting methods, where a model is trained per object, and 2) Generalizable models, which learn a universal 3D representation applicable across objects with minimal adaptation. Both approaches require high-quality input views and pose information, and a general formulation for both can be expressed as in Eq. (\ref{eq:eschernet2}). We experiment with both methods and propose a simple yet effective way to enhance a feed-forward generalizable model in a training-free manner.

\begin{equation} 
f(x) : \mathbb{R}^3 \to \mathbb{R}
\label{eq:eschernet2}
\end{equation}

\subsubsection{Overfitting Method}
The prior work EscherNet opts to train separate NeuS  ~\cite{wang2021neus} models for each object, which is able to memorize the details of a particular object by overfitting, leading to highly accurate and detailed reconstruction. 
For NeuS, $f(x)$ represents a learnable signed distance function (SDF) which maps a query from the 3D space to a signed distance to the closest object surface, with positive values outside and negative values inside the object. Such overfitting method can yield high-quality reconstruction however they usually involve extensive per-object training as shown in Sec. \ref{sec:exp3} which hinders implementation in time-pressured scenarios.

\begin{figure*}
  \centering
  \includegraphics[width=18cm]{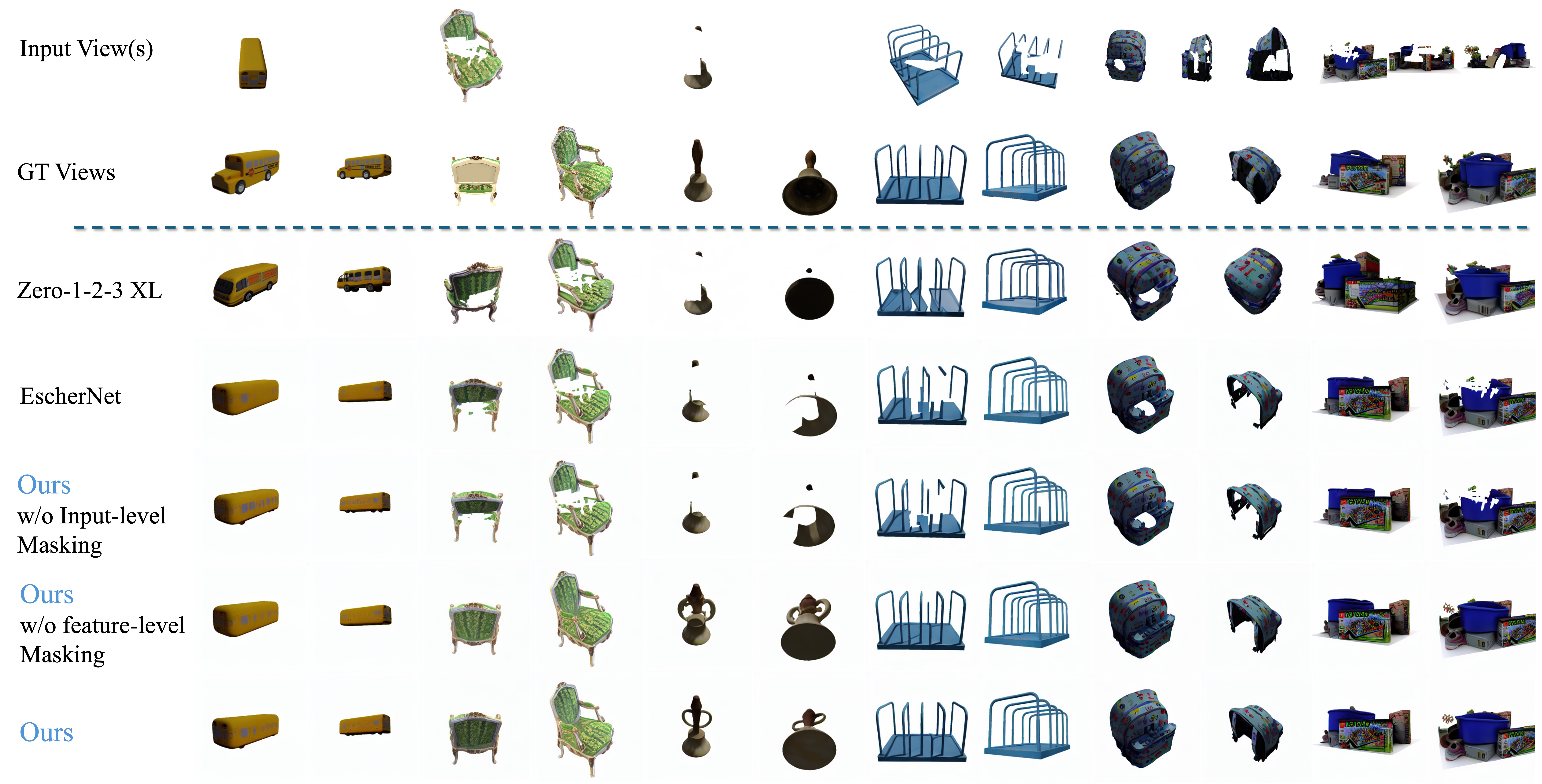}
  \caption{Visualization of synthesized views from different models with our OccNVS benchmark.}
  \label{fig:exp1}
\end{figure*}

\subsubsection{Generalizable Method}
There have been several feed-forward generalizable reconstruction models available in recent years  ~\cite{tang2023dreamgaussian, poole2022dreamfusion, hong2023lrm, tang2024lgm, xu2024instantmesh}, designed to quickly infer 3D representations from sparse inputs such as single view or a few views. %
For this group of models, $f(x)$ becomes a generalizable network which is able to provide the corresponding feature vectors which can be used to extract more explicit 3D representation according to the query point. However, despite their generalization capabilities, they are usually tailored for outputs from specific NVS models, which prohibits researchers from directly applying them with new NVS models. We pick one generalizable model, InstantMesh for case study in this paper. It is found InstantMesh performs worse when given inputs from poses other than those used in their paper, although their model design supports any input poses.

\textbf{Target View Synthesis:}  Luckily, we can take advantage of our model that can generate any view from any query pose, a feature inherited from EscherNet, to generate preferred views for generalizable reconstruction models. With the required views generated from our model, the performance of InstantMesh significantly surpasses that of straightforward integration. We further find that performance can be elevated if more generated views can be provided to the reconstruction model. No additional training of the reconstruction model is required in the whole process and no extra inference time is introduced as we show in Sec. \ref{sec:exp3} and App. ~\ref{app:exp2}.

\section{Experiments}
\label{sec:exp}

Experiments are conducted to compare our proposed method EscherNet++ with other state-of-the-art methods including our baseline EscherNet.

\textbf{Training Dataset:} Objaverse-1.0 is used to train our models. Specifically, a subset of 300K objects is sampled out of all 800K objects in the Objaverse-1.0 for faster fine-tuning and data-efficient purposes. We adopt the same rendered data used in Zero-1-to-3 and Eschernet. 
In our experiments, we set probability for image-level masking to be 0.5 for balanced learning on complete and masked input data, and vary the probability for feature-level masking to find the better setting, as detailed in Sec. ~\ref{sec:maskedtraining} and App. ~\ref{app:exp1}. It is found that 0.25 is a suitable probability for feature-level masking.

\textbf{Training Settings:} A small learning rate of ${1\cdot10^{-5}}$ is used for fine-tuning weights from the public checkpoint of EscherNet.
A batch size of 48 is adopted on each GPU, each training sample consisting of input views, target views of 256 × 256 resolution images and corresponding pose information for each object. 8 GPUs are used in training with the total batch size summed up to 384, which takes around 3 days to complete 28K iterations. The model checkpoint that achieves the highest validation performance is selected for testing. The structure of EscherNet++ and other settings are kept the same as EscherNet. 

\textbf{Test Settings \& Metrics:} 4DoF object-centric setting is set for all experiments. We evaluate all the models with two settings, one with complete input views and one with randomly occluded views with a new set of masks to simulate any possible occlusions from query viewpoints. 
We term the occluded benchmark {\bf OccNVS}, including complete/occluded views from Google Scanned Objects dataset (GSO) \cite{downs2022google}, RTMV and NeRF Synthetic \cite{mildenhall2021nerf}.

PSNR, SSIM \cite{wang2004image}, LPIPS \cite{zhang2018unreasonable} are adopted as metrics to evaluate the NVS and amodel completion performance of models. Chamfer Distance and Volume IoU are employed to measure the performance of 3D reconstruction. The other settings are alighed with Eshcernet. Implementation details can be found in App. \ref{app:exp2}.

\subsection{Results on Novel View Synthesis}
\label{sec:exp1}

\textbf{Overall Performance:} Our model and the base model EscherNet surpass Zero-1-2-3-based models in terms of overall performance in NVS without considering possible occlusions, as shown in Tab.~\ref{tab:exp1} and Fig.~\ref{fig:exp1} without occlusion. Powered by self and cross-attention to improve both target-to-target and reference-to-target consistency, EscherNet and our method can maintain more stable perspectives when synthesizing novel views. In contrary, Zero-1-2-3-based models can only synthesize one view at time and cannot reference multiple input views, restricting their abilities in optimizing multiple views coherently. 

\textbf{Robustness to Occlusions:} By masked fine-tuning, our model is designed to gain more robustness to possible occlusions in the input views. Experiments in Tab. \ref{tab:exp1} in occluded tests show that our model successfully achieves the intended goal of synthesizing complete novel views even with occlusion in input views, with semantic accuracy and geometric consistency well maintained. Therefore, it can be found that both qualitative and quantitative performance from our model in tasks with occlusion significantly outperforms baselines by improvement of at least 5 in PSNR for GSO in all settings over EscherNet.

\textbf{Semantic \& Geometry Understanding:}
Ablation studies on our hierarchical masked fine-tuning including input-level and feature-level masking are also conducted to examine the contributions from each component. Input-level masking plays the most important role to handle occlusions. The model achieves the ability to synthesize complete novel views after fine-tuning with input-level masking enabled by a curated and diverse dataset, as we compare the last three rows in Fig. \ref{fig:exp1}. Feature-level masking continue to contribute to understanding of semantics and geometry as shown in column 1, 7 and 10 in Fig. \ref{fig:exp1}. The combined hierarchical masking mechanism achieve the best overall performance in either regular or occluded tests as shown in Tab. \ref{tab:ab3}. It is found that the ratio for feature-level masking is crucial as we show in App. \ref{app:exp1}.

\begin{table*}

\centering

\resizebox{\textwidth}{!}{%

\begin{tabular}{lcccccccllllllllllll}
\toprule
\multirow{2}{*}{Method} & \multirow{2}{*}{\# Ref. Views} & \multicolumn{3}{c}{GSO-30} &\multicolumn{3}{c}{Occluded GSO-30} & \multicolumn{3}{c}{RTMV} & \multicolumn{3}{c}{Occluded RTMV} & \multicolumn{3}{c}{NeRF} & \multicolumn{3}{c}{Occluded NeRF}\\ 
\cmidrule(lr){3-5} \cmidrule(lr){6-8} \cmidrule(lr){9-11} \cmidrule(lr){12-14} \cmidrule(lr){15-17} \cmidrule(lr){18-20}
 &  & PSNR $\uparrow$ & SSIM $\uparrow$ & LPIPS $\downarrow$ &PSNR $\uparrow$ & SSIM $\uparrow$ & LPIPS $\downarrow$ & PSNR $\uparrow$ & SSIM $\uparrow$ &LPIPS $\downarrow$  & PSNR $\uparrow$ & SSIM $\uparrow$ &LPIPS $\downarrow$   & PSNR $\uparrow$ & SSIM $\uparrow$ &LPIPS $\downarrow$    & PSNR $\uparrow$ & SSIM $\uparrow$ &LPIPS $\downarrow$    \\

\midrule

Zero-1-to-3 \cite{liu2023zero} & 1& 18.55& 0.86& 0.122& 14.5\textcolor{gray}{+0.73}& 0.83\textcolor{gray}{+0.001}& 0.192\textcolor{gray}{-0.011}& 10.27& 0.514& 0.409& 9.33\textcolor{gray}{+0.16}& 0.505\textcolor{gray}{+0.002}& 0.428\textcolor{gray}{-0.003}& 12.61& 0.639& 0.31& 11.95\textcolor{gray}{-0.27}& 0.634\textcolor{gray}{-0.007}&0.338\textcolor{gray}{+0.01}\\
 Zero-1-to-3 XL \cite{liu2023zero} & 1& 18.74& 0.855& 0.124& 14.55\textcolor{gray}{+0.31}& 0.823\textcolor{gray}{-0.002}& 0.198\textcolor{gray}{-0.003}& 10.47& 0.516& 0.402& 9.38\textcolor{gray}{+0.27}& 0.503\textcolor{gray}{-0.001}& 0.429\textcolor{gray}{-0.005}& 12.62& 0.637& 0.309& 11.65\textcolor{gray}{-0.02}& 0.625\textcolor{gray}{-0.005}&0.346\textcolor{gray}{+0.001}\\ 
 
\midrule
\multirow{5}{*}{EscherNet\cite{kong2024eschernet}} & 1 & 20.05 &  0.883& 0.096 &15.64\textcolor{gray}{+0.99}& 0.852\textcolor{gray}{+0.004}& 0.161\textcolor{gray}{-0.012}& 10.43 & 0.520 & 0.411 & 9.63\textcolor{gray}{-0.01}& 0.511\textcolor{gray}{-0.004}& 0.432\textcolor{gray}{-0.004}& 13.35 & 0.658 & 0.293 & 12.55\textcolor{gray}{-0.17} & 0.654\textcolor{gray}{-0.007} &0.317\textcolor{gray}{+0.006} 
\\
 & 2 &  22.85&  0.908&  0.063 &15.82\textcolor{gray}{+1.29}& 0.865\textcolor{gray}{+0.003}& 0.145\textcolor{gray}{-0.016}& 12.55 & 0.581 & 0.306 & 10.92\textcolor{gray}{+0.16}& 0.566\textcolor{gray}{-0.005} & 0.344\textcolor{gray}{+0.003} 
& 14.93 & 0.699 & 0.210 & 13.39\textcolor{gray}{+0.42} & 0.685\textcolor{gray}{+0.003} &0.253\textcolor{gray}{-0.011} 
\\
 & 3 & 23.87 &  0.918& 0.052 &16.32\textcolor{gray}{+1.56}& 0.874\textcolor{gray}{+0.005}& 0.13\textcolor{gray}{-0.019}& 13.58 & 0.611 & 0.259 & 11.68\textcolor{gray}{+0.3} & 0.594\textcolor{gray}{+0.000} & 0.295\textcolor{gray}{-0.002} 
& 16.19 & 0.729 & 0.161 & 14.57\textcolor{gray}{+0.22} & 0.716\textcolor{gray}{+0.002} &0.119\textcolor{gray}{+0.072} 
\\
 & 5 & 24.91 &  0.926 & 0.044 &16.67\textcolor{gray}{+1.97}& 0.883\textcolor{gray}{+0.008}& 0.118\textcolor{gray}{-0.02}& 14.48 & 0.633 & 0.222 & 12.28\textcolor{gray}{+0.2} & 0.611\textcolor{gray}{+0.002} & 0.264\textcolor{gray}{-0.003} 
& 17.11 & 0.748 & 0.128 & 15.28\textcolor{gray}{+0.32} & 0.731\textcolor{gray}{+0.002} &0.167\textcolor{gray}{-0.006} 
\\
 & 10 & 25.65 & 0.933 & 0.037 &16.92\textcolor{gray}{+1.72}& 0.889\textcolor{gray}{+0.007} & 0.111\textcolor{gray}{-0.017}  & 15.44 & 0.657 & 0.186 & 13.00\textcolor{gray}{+0.21} & 0.634\textcolor{gray}{+0.001} & 0.23\textcolor{gray}{-0.002} & 17.72 & 0.760 & 0.115 & 15.8\textcolor{gray}{+0.34} & 0.746\textcolor{gray}{+0.003} &0.15\textcolor{gray}{-0.006} \\ 

\midrule

\multirow{5}{*}{Ours}& 1 &  20.11&  0.883&  0.094 & 19.72 &  0.879 &  0.103  & 10.5 & 0.523 & 0.408 & 10.34 & 0.52 & 0.416 
& 13.35 & 0.661 & 0.29 & 13.51 & 0.666 &0.29 
\\
& 2 &  22.83&  0.908&  0.062 & 21.86 &  0.902 &  0.07  & 12.57 & 0.583 & 0.303 & 12.32 & 0.577 & 0.316 
& 14.96 & 0.698 & 0.21 & 14.74 & 0.692 &0.221 
\\
& 3 &  24.02&  0.918&  0.051 & 23.22 &  0.913 &  0.056  & 13.45 & 0.608 & 0.262 & 13.29 & 0.603 & 0.269 
& 16.14 & 0.727 & 0.164 & 15.85 & 0.721 &0.174 
\\
& 5 &  25.15&  0.926 &  0.043 & 24.22 &  0.921 &  0.047  & 14.38 & 0.631 & 0.223 & 14.16 & 0.627 & 0.232 
& 16.97 & 0.745 & 0.132 & 16.79 & 0.74 &0.138 
\\
& 10 &  25.98&  0.934 &  0.036 & 25.06 &  0.929 &  0.04  & 15.42 & 0.658 & 0.186 & 15.13 & 0.652 & 0.196 & 17.72 & 0.759 & 0.115 & 17.49 & 0.755 &0.121 \\ 

\midrule

\multirow{3}{*}{Ours w/o} & 1 &  20.33&  0.886&  0.091 &15.78 & 0.856 & 0.158 & 10.59 & 0.531 & 0.399 & 9.64 & 0.519 & 0.42 & 13.35 & 0.657 & 0.292 & 12.8 & 0.659 &0.309 \\
& 2 & 22.7 & 0.907 &  0.063 &15.87 & 0.866 & 0.145 & 12.66 & 0.585 & 0.299 & 10.99 & 0.57 & 0.336 & 14.97 & 0.7 & 0.209 & 13.47 & 0.688 &0.251 \\
\multirow{2}{*}{Input-Level Masking}& 3 & 23.92 &  0.918&  0.051 &16.35 & 0.875 & 0.129 & 13.59 & 0.611 & 0.258 & 11.62 & 0.595 & 0.294 & 16.16 & 0.728 & 0.165 & 14.53 & 0.714 &0.203 \\
& 5 & 25.00 &  0.927 &  0.043 &16.66 & 0.883 & 0.118 & 14.41 & 0.632 & 0.223 & 12.21 & 0.612 & 0.266 & 17.0 & 0.745 & 0.131 & 15.24 & 0.73 &0.169 \\
& 10 &  25.91&  0.934 &  0.036 &17.02 & 0.891 & 0.11 & 15.3 & 0.655 & 0.189 & 12.88 & 0.632 & 0.234 & 17.53 & 0.756 & 0.119 & 15.76 & 0.744 &0.152 \\ 

\midrule

\multirow{3}{*}{Ours w/o} & 1 & 19.95 & 0.88 & 0.1 & 19.31 &  0.875 &  0.109 & 10.78 & 0.53 & 0.391 & 10.57 & 0.526 & 0.405 & 13.47 & 0.658 & 0.289 & 13.57 & 0.66 &0.295 \\
& 2 & 22.72 & 0.907 & 0.064 & 21.65 &  0.9 &  0.073 & 12.57 & 0.582 & 0.301 & 12.26 & 0.575 & 0.315 & 14.98 & 0.697 & 0.211 & 14.69 & 0.691 &0.226 \\
\multirow{2}{*}{Feature-Level Masking}& 3 &  23.93& 0.917 & 0.052 & 22.97 &  0.91 &  0.059 & 13.5 & 0.609 & 0.259 & 13.31 & 0.609 & 0.259 & 16.25 & 0.729 & 0.163 & 15.91 & 0.721 &0.175 \\
& 5 &  25.05& 0.926 &  0.043 & 23.98 &  0.919 &  0.049 & 14.37 & 0.63 & 0.223 & 14.12 & 0.624 & 0.233 & 17.22 & 0.749 & 0.128 & 16.86 & 0.742 &0.138 \\
& 10 & 25.85 &  0.934 & 0.037 & 24.77 &  0.927 &  0.042 & 15.38 & 0.658 & 0.185 & 15.08 & 0.65 & 0.195 & 17.7 & 0.76 & 0.116 & 17.43 & 0.754 &0.123 \\ 

\bottomrule
\end{tabular}
}
\caption{Performance comparison on GSO-30, RTMV, NeRF Synthetic datasets and occluded counterparts (OccNVS).
\textcolor{gray}{+/- Gray numbers} indicate the performance change resulting from applying separate amodal completion~\cite{ozguroglu2024pix2gestalt} to the input data prior to processing by the NVS models.
}
\label{tab:exp1}
\end{table*}

\subsection{Results on Amodel Completion}

\begin{figure}
  \centering
  \resizebox{0.5\textwidth}{!}{%
  \includegraphics[width=18cm]{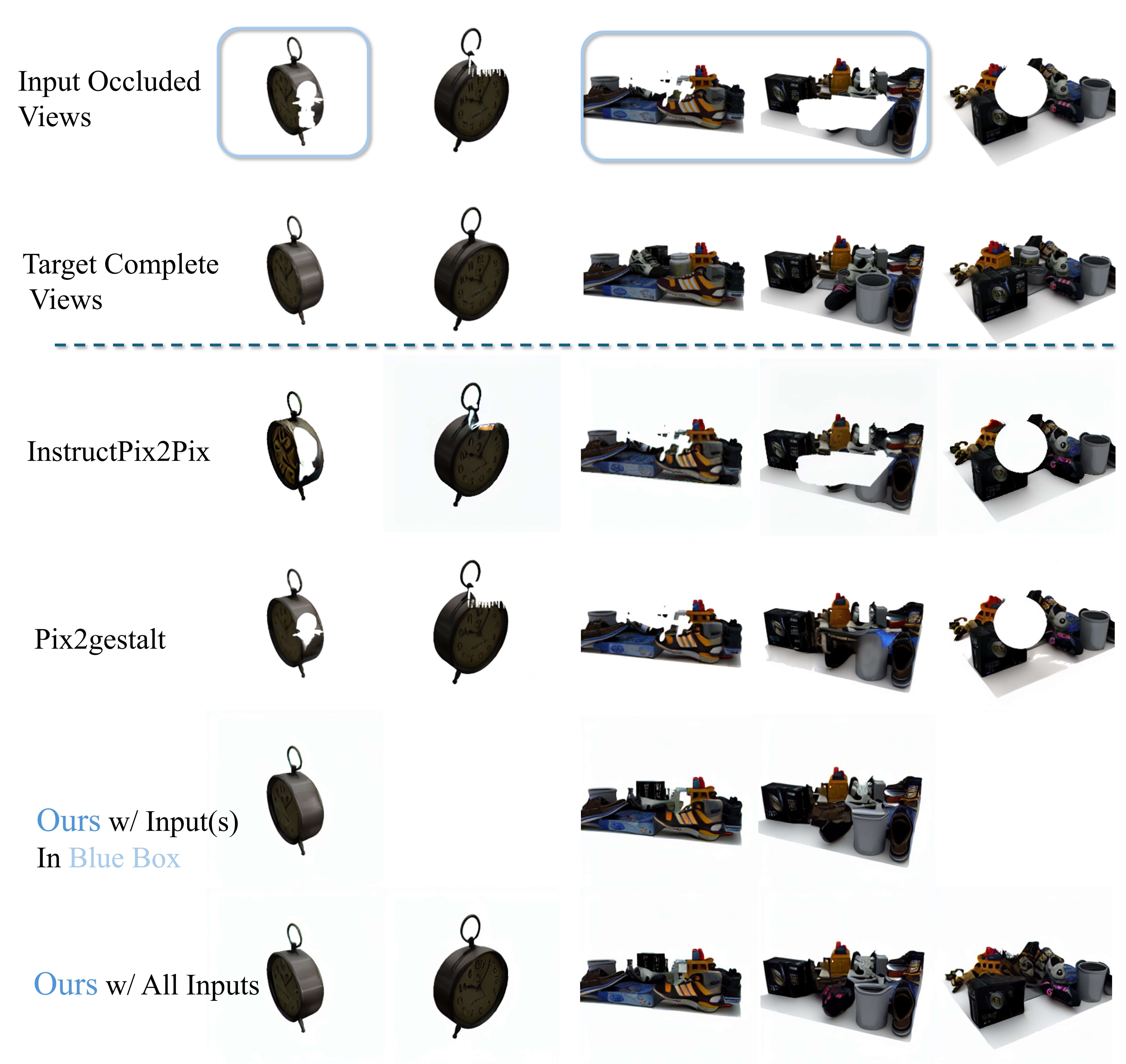}
  }
\caption{Amodal complation results by different models on OccNVS.}
\label{fig:exp2}
\end{figure}

We also compare the amodel completion performance of our model with two other recent models specifically designed for this task--InstructPix2Pix \cite{brooks2023instructpix2pix} trained in Gen3DSR \cite{Dogaru2024Gen3DSR} for amodel completion, and Pix2gestalt \cite{ozguroglu2024pix2gestalt}. 
For InstructPix2Pix and Pix2gestalt, one image is processed at a time by design, regardless of the number of occluded input views. For EscherNet++, the model takes in varying numbers of masked input images and synthesizes novel views of the object from different query viewpoints including the input viewpoints.

As shown in Fig. \ref{fig:exp2} and Tab. \ref{tab:exp2}, our model stands out for its ability to consider multi-view reference in amodel completion.

\begin{table}

\centering

\resizebox{0.5\textwidth}{!}{%
\begin{tabular}{lccccllllll}
\toprule
\multirow{2}{*}{Method} & \multirow{2}{*}{\# Ref. Views}  &\multicolumn{3}{c}{Occluded GSO-30}  & \multicolumn{3}{c}{Occluded RTMV}  & \multicolumn{3}{c}{Occluded NeRF}\\ 
\cmidrule(lr){3-5} \cmidrule(lr){6-8} \cmidrule(lr){9-11}
 &  &PSNR $\uparrow$ & SSIM $\uparrow$ & LPIPS $\downarrow$ & PSNR $\uparrow$ & SSIM $\uparrow$ &LPIPS $\downarrow$   & PSNR $\uparrow$ & SSIM $\uparrow$ &LPIPS $\downarrow$    \\

\midrule

\multirow{5}{*}{InstructPix2Pix \cite{brooks2023instructpix2pix, Dogaru2024Gen3DSR}}& 1& 18.08& 0.92& 0.098& 15.19& 0.829& 0.142& 16.77& 0.843&0.144\\
& 2& 17.84& 0.917& 0.107& 15.14& 0.837& 0.141& 17.49& 0.859&0.123\\
 & 3& 17.86& 0.918& 0.11& 15.03& 0.837& 0.141& 18.23& 0.868&0.117\\
 & 5& 17.88& 0.92& 0.108& 15.03& 0.824& 0.149& 18.71& 0.869&0.113\\
 & 10& 17.51& 0.918& 0.114& 15.38& 0.828& 0.145& 18.31& 0.872&0.111\\ 
 
\midrule
\multirow{5}{*}{Pix2gestalt \cite{ozguroglu2024pix2gestalt}}& 1 &20.71& 0.942& 0.072& 16.22& 0.85& 0.109& 16.98& 0.849&0.123\\
& 2 &19.87& 0.937& 0.082& 16.52& 0.859& 0.11& 17.45& 0.854&0.117\\
& 3 &20.2& 0.938& 0.08& 16.06& 0.859& 0.112& 18.02& 0.863&0.115\\
& 5 &20.38& 0.939& 0.079& 15.96& 0.852& 0.114& 18.43& 0.863&0.11\\
& 10 &19.94& 0.937& 0.084& 16.08& 0.85& 0.115& 18.2& 0.866&0.108\\ 

\midrule

\multirow{5}{*}{Ours} & 1 &28.42& 0.952& 0.029& 19.99& 0.832& 0.09& 21.24 & 0.841 &0.071\\
 & 2 &28.62& 0.954& 0.027& 20.93& 0.845& 0.078& 21.59& 0.852&0.065\\
 & 3 &29.29& 0.956& 0.025& 22.28& 0.848& 0.072& 22.22& 0.863&0.06\\
 & 5 &29.33& 0.957& 0.024& 22.26& 0.832& 0.075& 22.12& 0.858&0.06\\
 & 10 &28.34& 0.95& 0.027& 20.81& 0.799& 0.088& 21.47& 0.843&0.062\\ 

\bottomrule
\end{tabular}
}
\caption{Performance comparison on amodel completion on Occluded GSO-30, RTMV, and NeRF Synthetic datasets (OccNVS).
}
\label{tab:exp2}
\end{table}

\subsection{Results on 3D Reconstruction}
\label{sec:exp3}

\begin{figure}
  \centering
  \resizebox{0.5\textwidth}{!}{%
  \includegraphics[width=18cm]{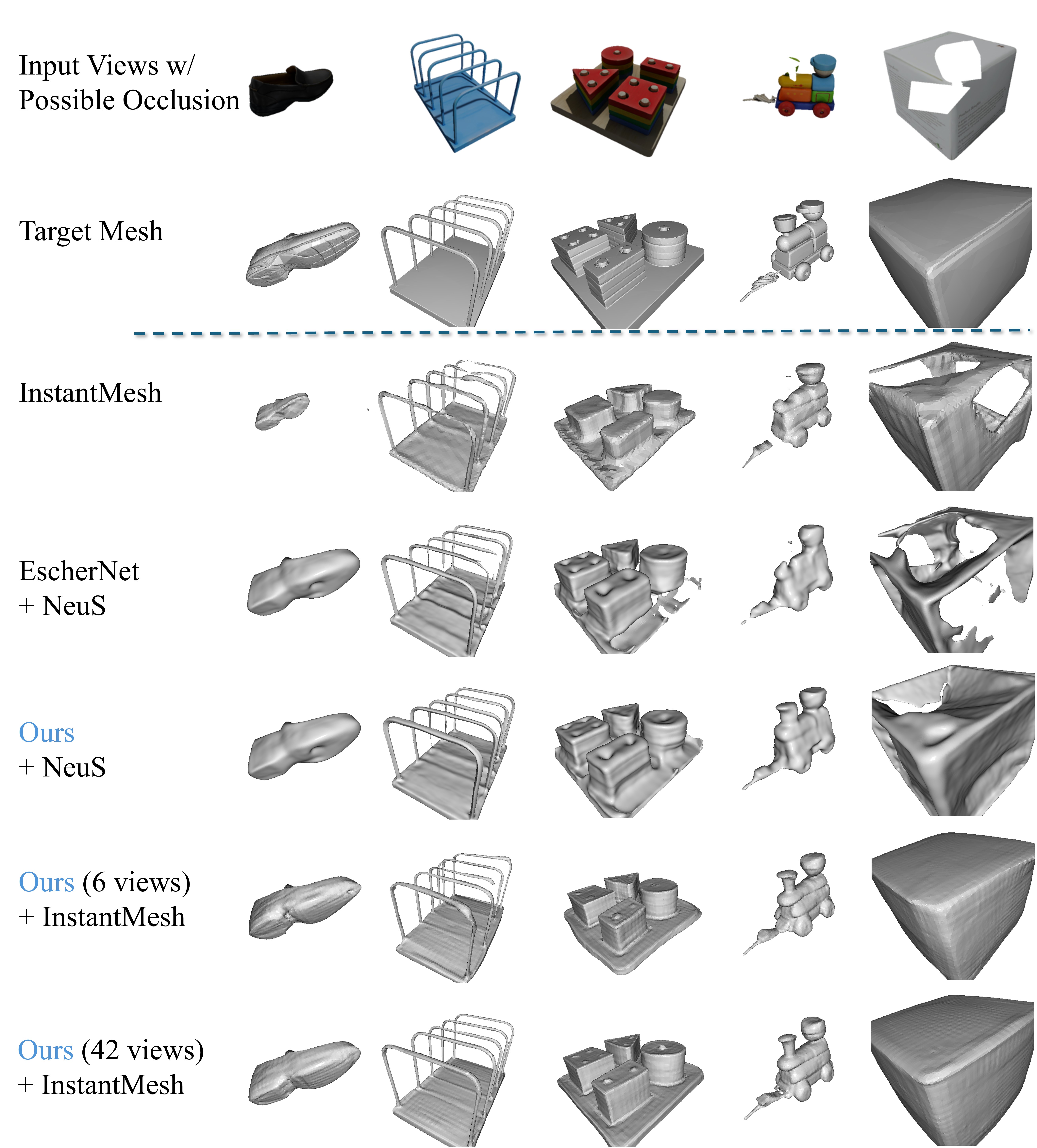}
  }
\caption{Rendered meshes from 3D reconstruction by different models on OccNVS benchmark. Note that a floater occurs in the first example with InstantMesh.}
\label{fig:exp3}
\end{figure}

We evaluate 3D reconstruction quality across various models, normalizing mesh outputs for comparison. Our model is benchmarked against recent state-of-the-art methods, including DreamGaussian \cite{tang2023dreamgaussian}, Large Multi-View Gaussian Model (LGM) \cite{tang2024lgm}, SyncDreamer \cite{liu2023syncdreamer}, InstantMesh \cite{xu2024instantmesh}, and EscherNet \cite{kong2024eschernet}. Additionally, we experiment with two image-to-3D reconstruction approaches: an overfitting method (NeuS \cite{wang2021neus}) and a feed-forward model (InstantMesh \cite{xu2024instantmesh}). Qualitative and quantitative results are presented in Fig.\ref{fig:exp3} and Tab.\ref{tab:exp3}. For our model, 36 synthesized views serve as inputs for NeuS-based reconstruction, while an additional 6 views are used for InstantMesh, totaling 42 views for enhanced reconstruction.

\textbf{Reconstruction by Overfitting:} Similar to NVS testing, our model and EscherNet outperform prior methods when paired with NeuS under no-occlusion settings (Fig.\ref{fig:exp3}, Tab.\ref{tab:exp3}). By synthesizing more consistent and precise views, they provide more accurate guidance for reconstruction. Notably, our model excels in handling occlusions in input views, leading to superior 3D reconstruction quality. However, this process requires iterative refinement, making it more computationally demanding than faster feed-forward methods (see Tab.~\ref{tab:exp3}).

\textbf{Enhanced Feed-Forward Reconstruction:} Existing feed-forward methods often underperform compared to overfitting approaches due to their reliance on NVS models for initial guidance, making their performance dependent on NVS quality. Artifacts like floaters from inconsistent input views degrade reconstruction, as shown in Fig.~\ref{fig:exp3} and App.~\ref{app:exp2}. However, our model synthesizes more consistent views at query viewpoints, enabling seamless integration with pre-trained feed-forward 3D reconstruction models. We validate this by integrating InstantMesh, achieving over a 10\% increase in volume IoU by providing more accurate views at the same viewpoints. Additionally, the reconstruction pipeline becomes more robust to occlusions, and performance further improves with increased view coverage. With this optimization, we reduce reconstruction time by 95\% while maintaining competitive performance, as shown in Tab.~\ref{tab:exp3} and Tab.~\ref{tab:exp4}.

\begin{table}

\resizebox{0.5\textwidth}{!}{%
\label{tab:3dexp}
\begin{tabular}{lclccccl}
\toprule
\multirow{2}{*}{Method} & \multirow{2}{*}{\# Ref. Views}  &\multirow{2}{*}{\# Nol. Views}& \multicolumn{2}{c}{GSO3D} & \multicolumn{2}{c}{Occluded GSO3D} & Time\\ 
\cmidrule(lr){4-5} \cmidrule(lr){6-7} \cmidrule(lr){8-8} 
 &   && Chamfer Dist. $\downarrow$ & Volume IoU $\uparrow$ & Chamfer Dist. $\downarrow$ & Volume IoU $\uparrow$   & Minutes $\downarrow$\\
 
 \midrule
 
 Dream Gaussian\cite{tang2023dreamgaussian}& 1& -& 0.0543& 0.4515& 0.0611& 0.3448& 1.5\\
 ImageDream\cite{wang2023imagedream}+LGM\cite{tang2024lgm}& 1& 4& 0.0877& 0.2521& 0.1787& 0.095& 1.5\\
SyncDreamer\cite{liu2023syncdreamer}+NeuS\cite{wang2021neus} & 1& 16& 0.0427& 0.5191& 0.0624& 0.2784& 27\\ 
Zero123++\cite{shi2023zero123++}+InstantMesh\cite{xu2024instantmesh} & 1  &6& 0.0608 & 0.4557 & 0.0655 & 0.2478   & 1.6\\ 
 
\midrule
\multirow{5}{*}{EscherNet \cite{kong2024eschernet} + NeuS\cite{wang2021neus}} & 1  &36& 0.0312 & 0.5941 & 0.0477 & 0.3736   & \multirow{5}{*}{27}\\
& 2  &36& 0.0217 & 0.6878 & 0.0671 & 0.286   & \\
& 3  &36& 0.0186 & 0.7117 & 0.0346 & 0.3853   & \\
& 5  &36& 0.0177 & 0.7377 & 0.0351 & 0.3976   & \\
& 10  &36& 0.0169 & 0.7442 & 0.0312 & 0.4498   & \\

\midrule

\multirow{5}{*}{Ours + NeuS} & 1  &36& 0.0305 & 0.6018 & 0.0376  & 0.5602   & \multirow{5}{*}{27}\\
& 2  &36& 0.0214 & 0.6921 & 0.0249  & 0.664   & \\
& 3  &36& 0.0185 & 0.7277 & 0.0197 & 0.7139   & \\
& 5  &36& 0.0182 & 0.7294 & 0.0189 & 0.7221  & \\
& 10  &36& 0.0168 & 0.7437 & 0.0176 & 0.7352   & \\ 

\midrule

\multirow{5}{*}{Ours + InstantMesh} & 1  &6& 0.0304 & 0.5912 & 0.0392  & 0.5405   & \multirow{5}{*}{1.3}\\
 & 2  &6& 0.0259 & 0.633 & 0.0301  & 0.5954   & \\
 & 3  &6& 0.0251 & 0.6491 & 0.0257 & 0.6413   & \\
 & 5  &6& 0.0238 & 0.6667 & 0.0291 & 0.6376  & \\
 & 10  &6& 0.0275 & 0.6472 & 0.0282 & 0.6414   & \\ 

\midrule

\multirow{5}{*}{Ours + InstantMesh} & 1  &42& 0.0278 & 0.6244 & 0.04  & 0.5501   & \multirow{5}{*}{1.3}\\
& 2  &42& 0.0224 & 0.6803 & 0.0311  & 0.6118   & \\
& 3  &42& 0.0265 & 0.6744 & 0.0277 & 0.6605   & \\
& 5  &42& 0.0253 & 0.6857 & 0.024 & 0.6886  & \\
& 10  &42& 0.0179 & 0.7295 & 0.0233 & 0.6987   & \\

\bottomrule
\end{tabular}
}
\caption{3D reconstruction comparison on GSO3D and Occluded GSO3D datasets. Time is measured from when input views are given to networks to when the reconstructed meshes are ready in the batch inference mode.
}
\label{tab:exp3}
\end{table}

\section{Conclusion}

In this paper, we propose EscherNet++, a masked fine-tuned diffusion model that can synthesize novel views of objects in a zero-shot way with amodal completion ability. We find that properly masked input images and input feature maps can contribute to better performance of the model. In addition, it can be seamlessly integrated with other fast feed-forward image-to-mesh models because of its flexible feature to synthesize any query views without the need for extra training, and the fast 3D reconstruction performance can be further boosted by its scalable nature. Limitations of the current work can be found in App.~\ref{app:limitation} as well as future work.

{
    \small
    \bibliographystyle{ieeenat_fullname}
    \bibliography{main}
}
\clearpage

\renewcommand{\thesection}{\Alph{section}} %
\renewcommand{\thesubsection}{\Alph{section}.\arabic{subsection}} %

\setcounter{page}{1}
\maketitlesupplementary

\appendix
\section{Ablation Study on Feature-Level Masking}
\label{app:exp1}

In experiment, we empirically find the proper ratio for feature-level masking. Consider a batch of feature maps from the image encoder, its tenser shape is $[b*t, l, c]$, in which $b$ is the batch size of samples, $t$ is number of input views in each sample, $l$ is the feature map area (number of feature vectors associated with each input view) and $c$ is the feature dimension.

We start by masking all (100\% of $b*t$ dimension) feature maps by half feature map area (50\% of $l$) randomly and the performance is sub-optimal. Then we gradually decrease the ratio on the second dimension by 25\% (in $b*t$ dimension), and finally found that 25 \% is a proper ratio for feature-level masking. That is, we report performance of the model with 25\% masked in $b*t$ dimension and 50\% masked in $l$ dimension in training, as the representative results of feature-level masking.

\begin{figure*}
  \centering
  \resizebox{1.0\textwidth}{!}{%
  \includegraphics[width=18cm]{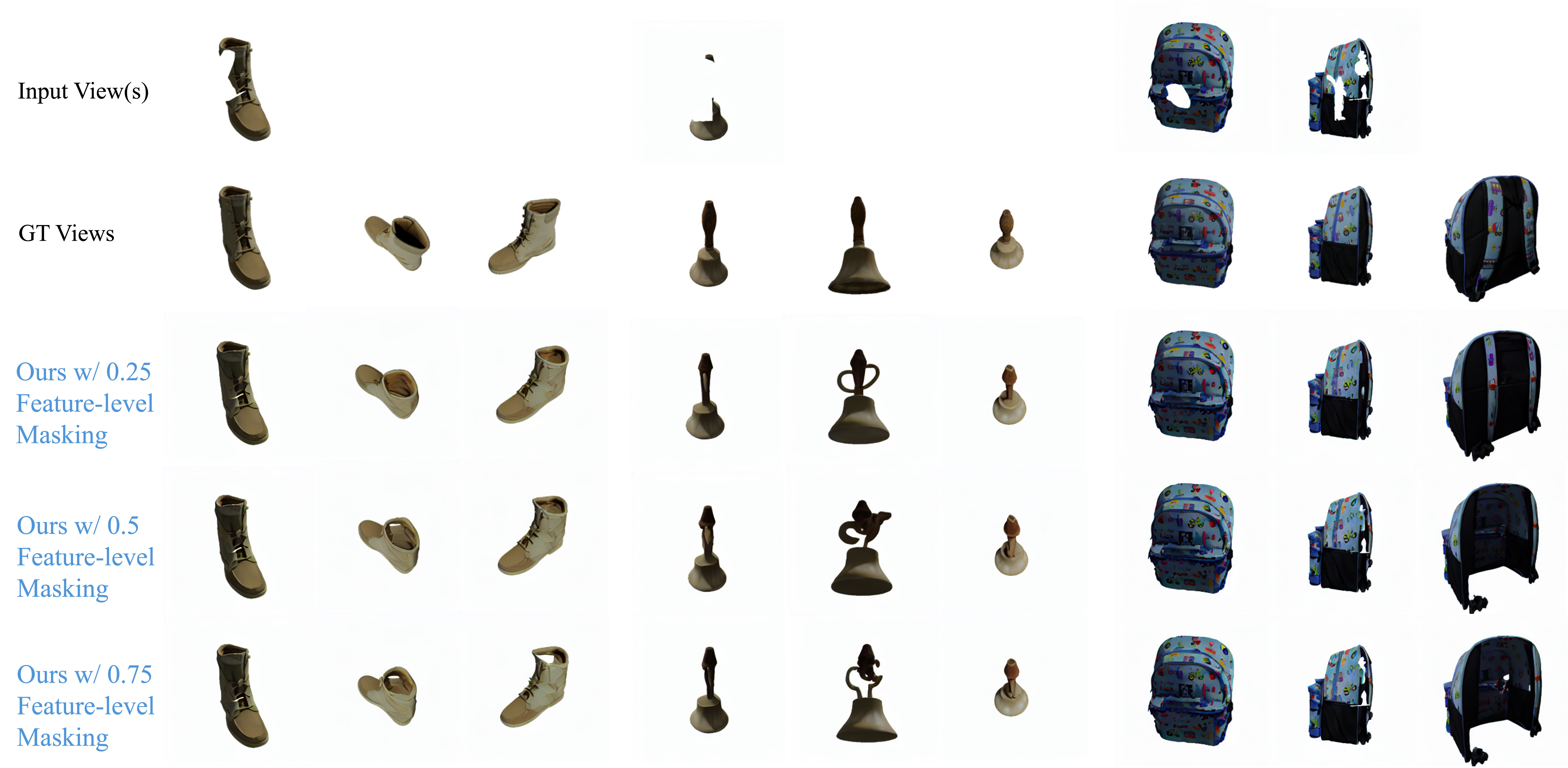}
  }
\caption{Qualitative results with different ratios for feature-level masking.}
\label{fig:exp5}
\end{figure*}

We also attach the full tables for evaluating models with {\bf OccNVS} in the ablation study on feature-level masking. It is found that feature-level masking with proper ratio can improve overall performance including better understanding of semantics from input views, better capture of intricate structures. However, it will lead to sub-optimal performance is too large ratio is picked, as shown is Fig. \ref{fig:exp5}, Tab. \ref{tab:ab1}, Tab. \ref{tab:ab2}, Tab. \ref{tab:ab3}.

\begin{table*}
\centering
\caption{Performance comparison on GSO-30 and Occluded GSO-30 datasets with different ratios for feature-level masking.}
\resizebox{\textwidth}{!}{%
\label{tab:ab1}
\begin{tabular}{lllccccccc}
\toprule
\multirow{2}{*}{Base Method}&\multirow{2}{*}{Input-Level Masking Ratio}&\multirow{2}{*}{Feature-Level Masking Ratio}& \multirow{2}{*}{\# Ref. Views} & \multicolumn{3}{c}{GSO-30} & \multicolumn{3}{c}{Occluded GSO-30} \\ 
\cmidrule(lr){5-7} \cmidrule(lr){8-10}
  & &&  & PSNR $\uparrow$ & SSIM $\uparrow$ & LPIPS $\downarrow$ & PSNR $\uparrow$ & SSIM $\uparrow$ & LPIPS $\downarrow$\\

\midrule

EscherNet (ckpt)& 0.5&1.0& 1 & 19.62& 0.879& 0.1& 19.11& 0.874& 0.11\\
EscherNet (ckpt)& 0.5&1.0& 2 & 22.21& 0.903& 0.067& 21.36& 0.897& 0.076\\
EscherNet (ckpt)& 0.5&1.0& 3 & 23.54& 0.915& 0.054& 22.58& 0.908& 0.061\\
EscherNet (ckpt)& 0.5&1.0& 5 & 24.51& 0.922& 0.046& 23.81& 0.917& 0.051\\
EscherNet (ckpt)& 0.5&1.0& 10 & 25.41& 0.93& 0.039& 24.68& 0.926& 0.043\\ 

\midrule

EscherNet (ckpt)& 0.5&0.75& 1 & 19.68& 0.879& 0.099& 19.21& 0.875& 0.108\\
EscherNet (ckpt)& 0.5&0.75& 2 & 22.4& 0.905& 0.066& 21.47& 0.898& 0.074\\
EscherNet (ckpt)& 0.5&0.75& 3 & 23.78& 0.916& 0.053& 22.65& 0.908& 0.061\\
EscherNet (ckpt)& 0.5&0.75& 5 & 24.82& 0.924& 0.044& 23.89& 0.918& 0.05\\
EscherNet (ckpt)& 0.5&0.75& 10 & 25.71& 0.933& 0.038& 24.84& 0.927& 0.042\\ 

\midrule

EscherNet (ckpt)& 0.5&0.5& 1 & 19.93& 0.883& 0.095& 19.27& 0.877& 0.107\\
EscherNet (ckpt)& 0.5&0.5& 2 & 22.72& 0.907 & 0.063& 21.76& 0.9& 0.072\\
EscherNet (ckpt)& 0.5&0.5& 3 & 23.87& 0.917& 0.051& 22.97& 0.91& 0.059\\
EscherNet (ckpt)& 0.5&0.5& 5 & 24.93& 0.925& 0.043& 24.04& 0.919& 0.049\\
EscherNet (ckpt)& 0.5&0.5& 10 & 25.88& 0.933 & 0.037 & 24.95& 0.927 & 0.041\\ 

\midrule

EscherNet (ckpt)& 0.5& 0.25& 1 & 20.11 & 0.883 & 0.094 & 19.72 & 0.879 & 0.103 \\
EscherNet (ckpt)& 0.5& 0.25& 2 & 22.83 & 0.908 & 0.062 & 21.86 & 0.902 & 0.07 \\
EscherNet (ckpt)& 0.5& 0.25& 3 & 24.02 & 0.918 & 0.051 & 23.22 & 0.913 & 0.056 \\
EscherNet (ckpt)& 0.5& 0.25& 5 & 25.15 & 0.926 & 0.043 & 24.22 & 0.921 & 0.047 \\
EscherNet (ckpt)& 0.5& 0.25& 10 & 25.98 & 0.934 & 0.036 & 25.06 & 0.929 & 0.04 \\ 

\midrule

EscherNet (ckpt)& 0.5&0& 1 & 19.95 & 0.88 & 0.1 & 19.31 & 0.875 & 0.109 \\
EscherNet (ckpt)& 0.5&0& 2 & 22.72 & 0.907 & 0.064 & 21.65 & 0.9 & 0.073 \\
EscherNet (ckpt)& 0.5&0& 3 & 23.93 & 0.917 & 0.052 & 22.97 & 0.91 & 0.059 \\
EscherNet (ckpt)& 0.5&0& 5 & 25.05 & 0.926 & 0.043 & 23.98 & 0.919 & 0.049 \\
EscherNet (ckpt)& 0.5&0& 10 & 25.85 & 0.934 & 0.037 & 24.77 & 0.927 & 0.042 \\ 

\bottomrule
\end{tabular}
}
\end{table*}

\begin{table*}
\centering
\caption{Performance comparison on RTMV and Occluded RTMV datasets with different ratios for feature-level masking.}
\resizebox{\textwidth}{!}{%
\label{tab:ab2}
\begin{tabular}{lllccccccc}
\toprule
\multirow{2}{*}{Base Method}&\multirow{2}{*}{Input-Level Masking Ratio}&\multirow{2}{*}{Feature-Level Masking Ratio}& \multirow{2}{*}{\# Ref. Views} & \multicolumn{3}{c}{RTMV} & \multicolumn{3}{c}{Occluded RTMV} \\ 
\cmidrule(lr){5-7} \cmidrule(lr){8-10}
  & &&  & PSNR $\uparrow$ & SSIM $\uparrow$ & LPIPS $\downarrow$ & PSNR $\uparrow$ & SSIM $\uparrow$ & LPIPS $\downarrow$\\

\midrule

EscherNet (ckpt)& 0.5&1.0& 1 & 10.62& 0.532& 0.401& 10.37& 0.525& 0.414\\
EscherNet (ckpt)& 0.5&1.0& 2 & 12.38& 0.58& 0.31& 12.14& 0.574& 0.322\\
EscherNet (ckpt)& 0.5&1.0& 3 & 13.23& 0.606& 0.267& 13.02& 0.6& 0.279\\
EscherNet (ckpt)& 0.5&1.0& 5 & 14.23& 0.628& 0.232& 13.94& 0.62& 0.243\\
EscherNet (ckpt)& 0.5&1.0& 10 & 15.2& 0.654& 0.192& 14.96& 0.648& 0.201\\ 

\midrule

EscherNet (ckpt)& 0.5&0.75& 1 & 10.29& 0.522& 0.418& 10.12& 0.518& 0.428\\
EscherNet (ckpt)& 0.5&0.75& 2 & 12.3& 0.577& 0.316& 12.17& 0.576& 0.32\\
EscherNet (ckpt)& 0.5&0.75& 3 & 13.3& 0.606 & 0.267& 13.1& 0.6& 0.278 \\
EscherNet (ckpt)& 0.5&0.75& 5 & 14.3& 0.63& 0.227& 14.01& 0.623& 0.239\\
EscherNet (ckpt)& 0.5&0.75& 10 & 15.17& 0.652& 0.193& 14.9& 0.647& 0.203\\ 

\midrule

EscherNet (ckpt)& 0.5&0.5& 1 & 10.37& 0.521& 0.415 & 10.23 & 0.518 & 0.42 \\
EscherNet (ckpt)& 0.5&0.5& 2 & 12.3 & 0.575 & 0.318 & 12.08 & 0.571 & 0.327 \\
EscherNet (ckpt)& 0.5&0.5& 3 & 13.23 & 0.604 & 0.272 & 13.1 & 0.599 & 0.28 \\
EscherNet (ckpt)& 0.5&0.5& 5 & 14.26 & 0.628 & 0.229 & 14.02 & 0.622 & 0.24 \\
EscherNet (ckpt)& 0.5&0.5& 10 & 15.21 & 0.652 & 0.192 & 14.93 & 0.645 & 0.202 \\ 

\midrule

EscherNet (ckpt)& 0.5& 0.25& 1 & 10.5 & 0.523 & 0.408 & 10.34 & 0.52 & 0.416 \\
EscherNet (ckpt)& 0.5& 0.25& 2 & 12.57 & 0.583 & 0.303 & 12.32 & 0.577 & 0.316 \\
EscherNet (ckpt)& 0.5& 0.25& 3 & 13.45 & 0.608 & 0.262 & 13.29 & 0.603 & 0.269 \\
EscherNet (ckpt)& 0.5& 0.25& 5 & 14.38 & 0.631 & 0.223 & 14.16 & 0.627 & 0.232 \\
EscherNet (ckpt)& 0.5& 0.25& 10 & 15.42 & 0.658 & 0.186 & 15.13 & 0.652 & 0.196 \\ 

\midrule

EscherNet (ckpt)& 0.5&0& 1 & 10.78 & 0.53 & 0.391 & 10.57 & 0.526 & 0.405 \\
EscherNet (ckpt)& 0.5&0& 2 & 12.57 & 0.582 & 0.301 & 12.26 & 0.575 & 0.315 \\
EscherNet (ckpt)& 0.5&0& 3 & 13.5 & 0.609 & 0.259 & 13.31 & 0.609 & 0.259 \\
EscherNet (ckpt)& 0.5&0& 5 & 14.37 & 0.63 & 0.223 & 14.12 & 0.624 & 0.233 \\
EscherNet (ckpt)& 0.5&0& 10 & 15.38 & 0.658 & 0.185 & 15.08 & 0.65 & 0.195 \\ 

\bottomrule
\end{tabular}
}
\end{table*}

\begin{table*}
\centering
\caption{Performance comparison on NeRF and Occluded NeRF datasets with different ratios for feature-level masking.}
\resizebox{\textwidth}{!}{%
\label{tab:ab3}
\begin{tabular}{lllccccccc}
\toprule
\multirow{2}{*}{Base Method}&\multirow{2}{*}{Input-Level Masking Ratio}&\multirow{2}{*}{Feature-Level Masking Ratio}& \multirow{2}{*}{\# Ref. Views} & \multicolumn{3}{c}{NeRF} & \multicolumn{3}{c}{Occluded NeRF} \\ 
\cmidrule(lr){5-7} \cmidrule(lr){8-10}
  & &&  & PSNR $\uparrow$ & SSIM $\uparrow$ & LPIPS $\downarrow$ & PSNR $\uparrow$ & SSIM $\uparrow$ & LPIPS $\downarrow$\\

\midrule

EscherNet (ckpt)& 0.5&1.0& 1 & 13.43& 0.657& 0.292& 13.5& 0.661& 0.295\\
EscherNet (ckpt)& 0.5&1.0& 2 & 14.99& 0.696& 0.214& 14.72& 0.688& 0.229\\
EscherNet (ckpt)& 0.5&1.0& 3 & 16.19& 0.728& 0.166& 15.87& 0.722& 0.178\\
EscherNet (ckpt)& 0.5&1.0& 5 & 17.01& 0.744& 0.133& 16.71& 0.738& 0.143\\
EscherNet (ckpt)& 0.5&1.0& 10 & 17.46& 0.754& 0.121& 17.19& 0.749& 0.128\\ 

\midrule

EscherNet (ckpt)& 0.5&0.75& 1 & 13.37& 0.659& 0.3& 13.9& 0.671& 0.282\\
EscherNet (ckpt)& 0.5&0.75& 2 & 14.93& 0.695& 0.214& 14.66& 0.688& 0.229\\
EscherNet (ckpt)& 0.5&0.75& 3 & 16.19& 0.727& 0.166& 15.87& 0.721& 0.177\\
EscherNet (ckpt)& 0.5&0.75& 5 & 17.12& 0.747& 0.13& 16.74& 0.739& 0.141\\
EscherNet (ckpt)& 0.5&0.75& 10 & 17.53& 0.756& 0.119& 17.26& 0.751& 0.126\\ 

\midrule

EscherNet (ckpt)& 0.5&0.5& 1 & 13.43 & 0.659 & 0.295 & 13.47 & 0.659 & 0.3 \\
EscherNet (ckpt)& 0.5&0.5& 2 & 14.85 & 0.695 & 0.212 & 14.66 & 0.689 & 0.224 \\
EscherNet (ckpt)& 0.5&0.5& 3 & 16.14 & 0.727 & 0.164 & 15.84 & 0.721 & 0.176 \\
EscherNet (ckpt)& 0.5&0.5& 5 & 16.97 & 0.745 & 0.132 & 16.69 & 0.738 & 0.142 \\
EscherNet (ckpt)& 0.5&0.5& 10 & 17.4 & 0.754 & 0.121 & 17.16 & 0.749 & 0.128 \\ 

\midrule

EscherNet (ckpt)& 0.5& 0.25& 1 & 13.35 & 0.661 & 0.29 & 13.51 & 0.666 & 0.29 \\
EscherNet (ckpt)& 0.5& 0.25& 2 & 14.96 & 0.698 & 0.21 & 14.74 & 0.692 & 0.221 \\
EscherNet (ckpt)& 0.5& 0.25& 3 & 16.14 & 0.727 & 0.164 & 15.85 & 0.721 & 0.174 \\
EscherNet (ckpt)& 0.5& 0.25& 5 & 16.97 & 0.745 & 0.132 & 16.79 & 0.74 & 0.138 \\
EscherNet (ckpt)& 0.5& 0.25& 10 & 17.72 & 0.759 & 0.115 & 17.49 & 0.755 & 0.121 \\ 

\midrule

EscherNet (ckpt)& 0.5&0& 1 & 13.47 & 0.658 & 0.289 & 13.57 & 0.66 & 0.295 \\
EscherNet (ckpt)& 0.5&0& 2 & 14.98 & 0.697 & 0.211 & 14.69 & 0.691 & 0.226 \\
EscherNet (ckpt)& 0.5&0& 3 & 16.25 & 0.729 & 0.163 & 15.91 & 0.721 & 0.175 \\
EscherNet (ckpt)& 0.5&0& 5 & 17.22 & 0.749 & 0.128 & 16.86 & 0.742 & 0.138 \\
EscherNet (ckpt)& 0.5&0& 10 & 17.7 & 0.76 & 0.116 & 17.43 & 0.754 & 0.123 \\ 

\bottomrule
\end{tabular}
}
\end{table*}

\section{Implementation Details of Models in Comparison}
\label{app:exp2}

We compare our model with several recent SoTA models: Zero-1-2-3, Zero-1-2-3 XL \cite{liu2023zero} and EscherNet \cite{kong2024eschernet} for comparison in NVS tasks; DreamGaussian \cite{tang2023dreamgaussian}, Large Multi-View Gaussian Model(LGM) \cite{tang2024lgm}, SyncDreamer \cite{liu2023syncdreamer}, InstantMesh \cite{xu2024instantmesh} and EscherNet \cite{kong2024eschernet} for mesh quality comparison in 3D reconstruction tasks. OccNVS is used for comparison. For 3D reconstruction tasks, raw meshes from the models are normalized first and then compared with ground truth as in \cite{liu2023syncdreamer, kong2024eschernet}.

\textbf{Zero-1-2-3 \& Zero-1-2-3 XL} It is the first work in diffusion-based NVS for objects. In its model design, one input view can be referenced at a time and one target view can be synthesized afterwards. As a result, Zero-1-2-3 and its XL version are only adopted for one-input settings.

\textbf{EscherNet} Our model shares the same model structure with EscherNet. As the result, EscherNet can be used for direct comparison in all tasks and settings in this paper, including NVS and 3D reconstruction. For NVS, EscherNet is able to synthesize multiple novels view from any query viewpoints. For 3D reconstruction, 36 fixed view are synthesized, with the azimuth from 0\degree to 360\degree with a rendering every 30\degree at a set of elevations (-30\degree, 0\degree, 30\degree) for reconstruction with NeuS, the same setting as reconstruction with our model.

We fine-tune our model based on public weights shared by authors of Eschernet, and we have confirmed with them about the performance of EscherNet in the experiments.

\textbf{DreamGaussian} It is a two-stage model, which uses the first stage for reconstruction conditioned on a single input view and second image for texture refinement. Hence, there are no novel views required before reconstruction. Rotation is conducted for evaluation as in EscherNet. It is worth noting that DreamGaussian and LGM are the fastest methods for reconstruction in our experiment.

\begin{figure}
  \centering
  \resizebox{0.5\textwidth}{!}{%
  \includegraphics[width=15cm]{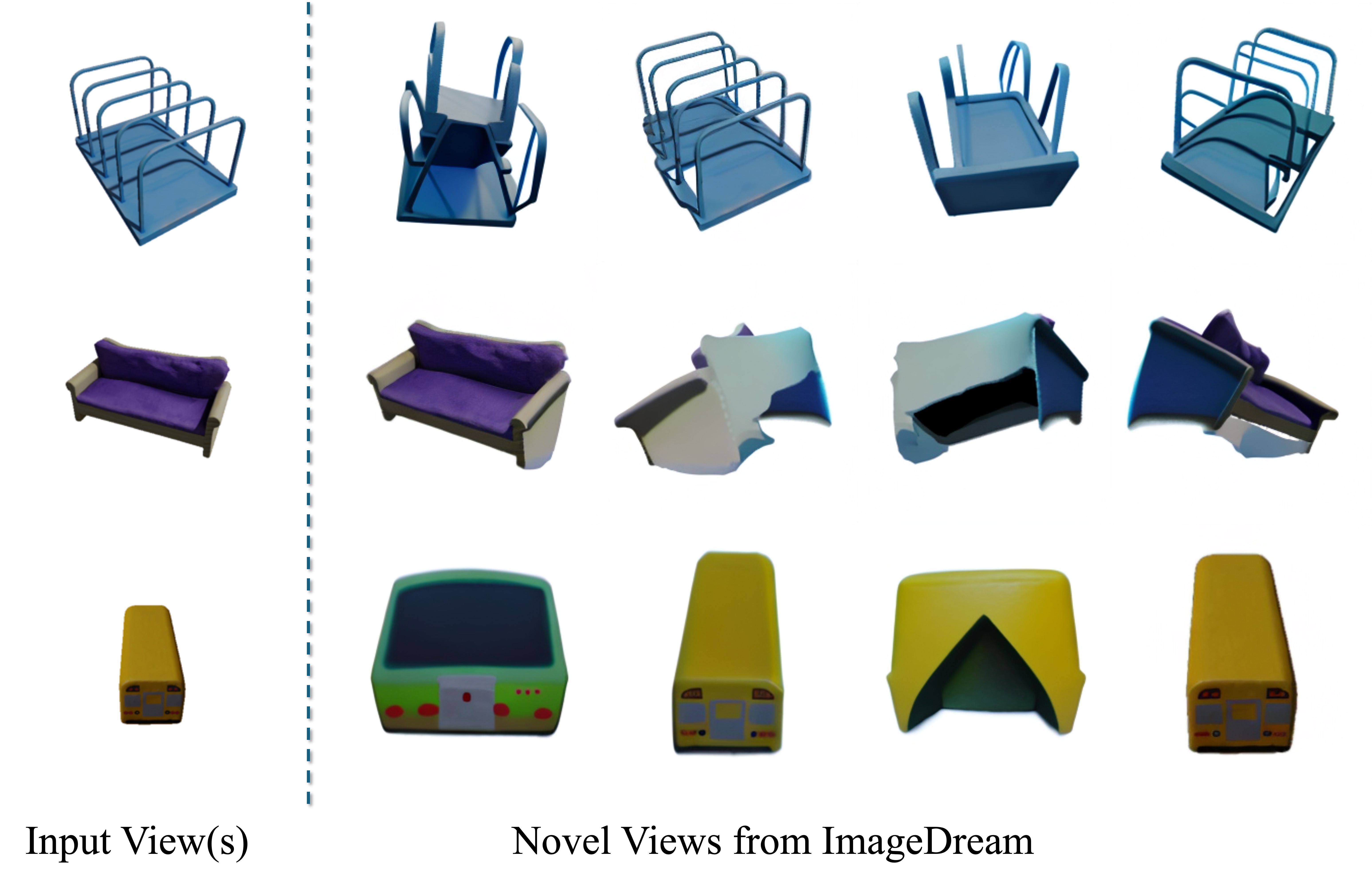}
  }
\caption{Examples of novel views generated by ImageDream. It struggles with significant elevations and azimuths. Therefore, the challenge is propagated to the reconstruction pipeline of LGM.}
\label{fig:exp6}
\end{figure}

\textbf{LGM}  As a two-stage method, LGM \cite{tang2024lgm} depends on four views from fixed viewpoints synthesized by ImageDream \cite{wang2023imagedream} conditioned on one input view to reconstruct 3D. It is also a fast pipeline, however, it is found to struggle with significant elevation and azimuth angles in input views. Therefore, it does not perform well in our tests. The fundamental reason is that ImageDream may not be able to provide consistent and reasonable novel views when conditioned on inputs with significant angles, as shown in Fig.\ref{fig:exp6}. The same rotation mechanism is conducted as with DreamGaussain.

\textbf{SyncDreamer} 16 fixes views are synthesized conditioned on one input view and then given to NeuS \cite{wang2021neus} by SyncDreamer \cite{liu2023syncdreamer}. Compared with reconstruction time which usually takes near 30 minutes, the time spent on synthesis is almost insignificant. That is, the time used to reconstruct an object from one input view to a complete mesh is largely dependent on the reconstruction method, which shares a similar case with reconstitution based on our model with overfitting methods like NeuS. 

\begin{figure}
  \centering
  \resizebox{0.5\textwidth}{!}{%
  \includegraphics[width=15cm]{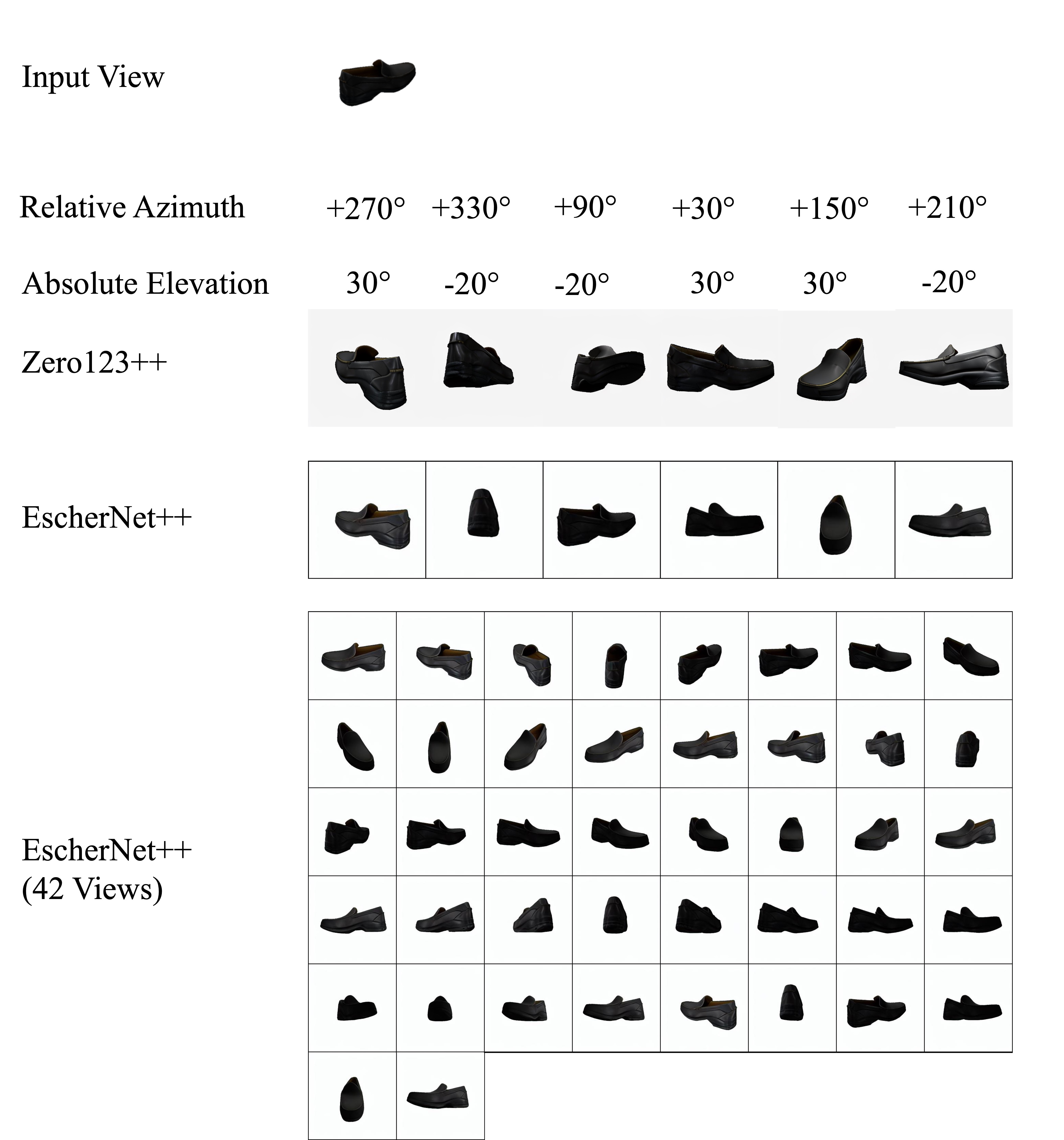}
  }
\caption{Examples of novel views generated by Zero123++ and EscherNet++ for reconstruction by InstantMesh. The last row contains all 42 views by our model. The scale and pose of the object in novel views by Zero123++ are not consistent sometimes, which can lead to confusion for InstantMesh.}
\label{fig:exp7}
\end{figure}

\textbf{InstantMesh} In the original pipeline, Zero123++ \citet{shi2023zero123++} is used for NVS at the first stage and InstantMesh \cite{xu2024instantmesh} construct the mesh based on novel views. Zero123++ is designed to generate 6 fixed views of an object with relative azimuth rotations and absolute elevations.  The 6 input images have poses with alternating absolute elevations of 20\textdegree  and -10\textdegree, and their azimuths are defined relative to the query image, beginning at 30\textdegree and increased by 60\textdegree for subsequent poses. However, it sometimes generate meshes with floaters around the object, which leads to erroneous scale in normalization, as shown in Fig. \ref{fig:exp3}. It is found that we can make use of our model to generate more consistent novel views at the preferred viewpoints for InstantMesh so that the performance can be improved significantly without floaters in the final meshes. The performance can be further enhanced by providing more novel views covering more viewpoints to InstantMesh. We provide one example comparing novel views from Zero123++ and our method in Fig. \ref{fig:exp7}. No extra training or extra reference time is induced in this whole process.

Although it is able to provide views from any viewpoints, we find that the six viewpoints used in the original pipeline and their absolute values are necessary to the network. Therefore, we define that the input views are at 0\degree azimuth angle and we rotate the meshes back before evaluation.

As noticed by authors of InstantMesh, InstantMesh is able to take in various numbers of input views because of its transformer-based structure. However, in contrast to their finding that decrease the number of input views can boost the performance in some hard cases, we found with our model, simply increasing the number of input views can further improve the overall reconstruction performance without extra overheads, thanks to the ability to synthesize high-quality views from any query viewpoints from our model.

\section{Limitations \& Future Work} 
\label{app:limitation}

\begin{figure}
  \centering
  \resizebox{0.5\textwidth}{!}{%
  \includegraphics[width=15cm]{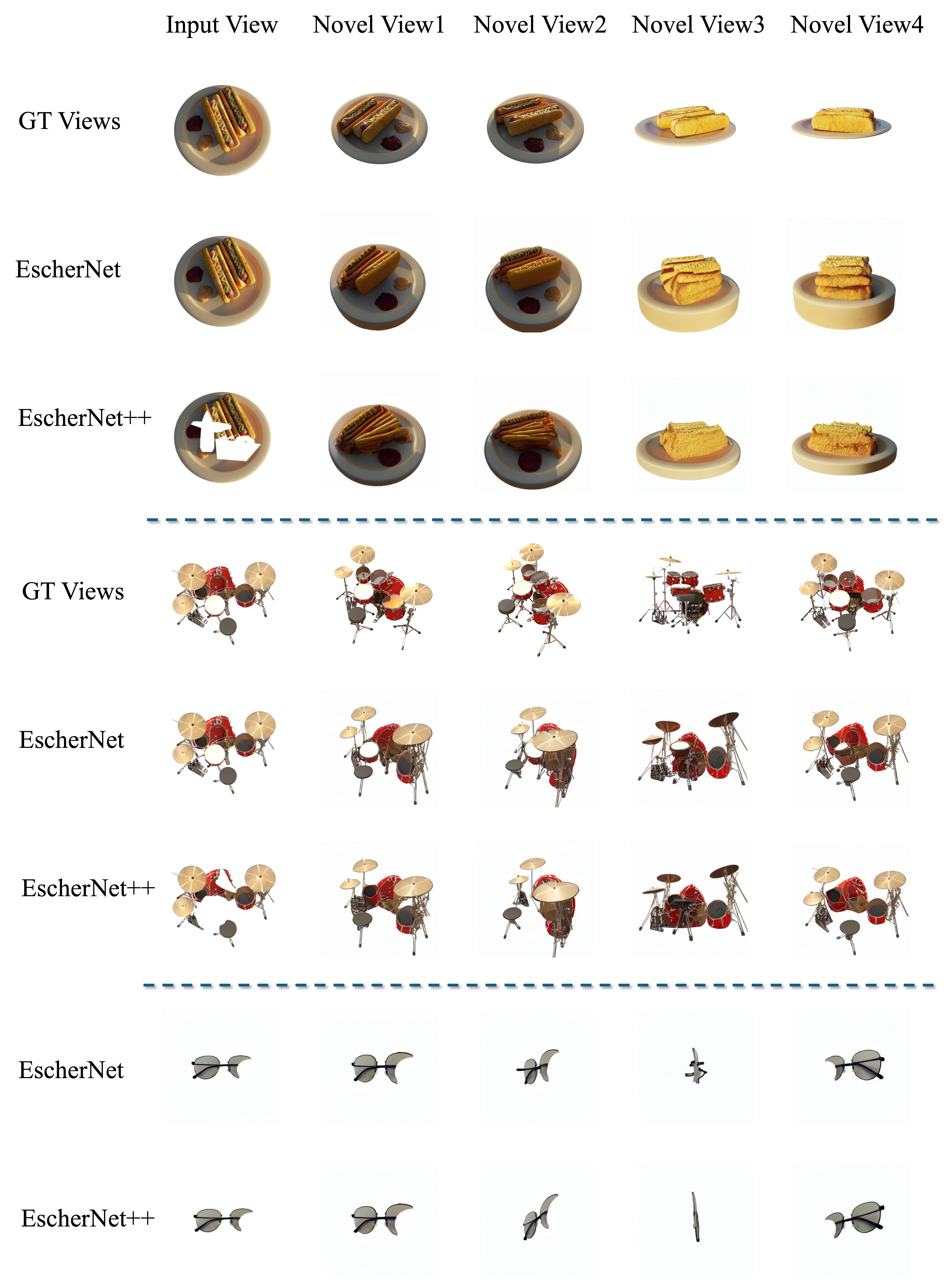}
  }
\caption{Examples of typical error cases of EscherNet++.}
\label{fig:error}
\end{figure}

During our experiments, we identified two notable limitations in our model’s performance. 

1) EscherNet++ exhibits degraded performance on inputs that contain intricate details and complex spatial layouts, similar to the performance of the base model EscherNet. This is evident in the first two rows of Fig.~\ref{fig:error}, where despite the model’s ability to infer novel views and fill in occluded regions, the synthesized outputs lack both visual and semantic consistency. Fine-grained textures and structural elements are either oversimplified or inconsistently rendered across views, suggesting that the models struggle to maintain coherence when completing regions with high-frequency details or complex shapes, partially because of the limited resolution (256*256) of both input and output views.  2) It possibly presents failure when presented with out-of-distribution (OOD) occluded inputs. The third row in Fig.~\ref{fig:error} provides an example: the input includes a partially visible pair of eyeglasses, yet the model fails to recognize the object's class or underlying geometry. Instead of completing the missing regions in a plausible way, it generates a flat, unstructured form that resembles a piece of paper. The underlying reason is the lack of understanding in OOD inputs.

Future work can explore robust architecture designs with more diverse datasets, more explicit guidance with multi-modal inputs, including more expressive visual and semantic features. Feed-forward 3D reconstruction methods also have the potential to be improved in terms of how to increase robustness to inconsistency in inputs views and utilize increasing number of views more efficiently. Last, a comprehensive framework is necessary to make our work more accessible in applications that includes object segmentation, pose estimation, etc, combined as integrated modules or a single unified model.

\begin{table}

\resizebox{0.5\textwidth}{!}{%
\label{tab:3dexp}
\begin{tabular}{lclccccl}
\toprule
\multirow{2}{*}{Method} & \multirow{2}{*}{\# Ref. Views}  &\multirow{2}{*}{\# Nol. Views}& \multicolumn{2}{c}{GSO3D} & \multicolumn{2}{c}{Occluded GSO3D} & Time\\ 
\cmidrule(lr){4-5} \cmidrule(lr){6-7} \cmidrule(lr){8-8} 
 &   && Chamfer Dist. $\downarrow$ & Volume IoU $\uparrow$ & Chamfer Dist. $\downarrow$ & Volume IoU $\uparrow$   & Minutes $\downarrow$\\
 
 \midrule
 Zero123++\cite{shi2023zero123++}+InstantMesh\cite{xu2024instantmesh} & 1  &6& 0.0608 & 0.4557 & 0.0655 & 0.2478   & 1.6\\ 
 ImageDream\cite{wang2023imagedream}+LGM\cite{tang2024lgm}& 1& 4& 0.0877& 0.2521& 0.1787& 0.095& 1.5\\
 
\midrule

\multirow{5}{*}{Ours + InstantMesh} & 1  &6& 0.0304 & 0.5912 & 0.0392  & 0.5405   & \multirow{5}{*}{1.3}\\
 & 2  &6& 0.0259 & 0.633 & 0.0301  & 0.5954   & \\
 & 3  &6& 0.0251 & 0.6491 & 0.0257 & 0.6413   & \\
 & 5  &6& 0.0238 & 0.6667 & 0.0291 & 0.6376  & \\
 & 10  &6& 0.0275 & 0.6472 & 0.0282 & 0.6414   & \\ 

\midrule

\multirow{5}{*}{Ours + LGM} & 1  &4& 0.0337& 0.57& 0.0414& 0.5099& \multirow{5}{*}{1.5}\\
 & 2  &4& 0.0293& 0.613& 0.031& 0.5847& \\
 & 3  &4& 0.0269& 0.6247& 0.0283& 0.6223& \\
 & 5  &4& 0.0256& 0.6242& 0.0266& 0.6127& \\
 & 10  &4& 0.0248& 0.6471& 0.0264& 0.6345& \\

\bottomrule
\end{tabular}
}
\caption{3D reconstruction comparison from InstantMesh and LGM on GSO3D and Occluded GSO3D datasets, with different integration with NVS models.
}
\label{tab:exp4}
\end{table}

\end{document}